\documentclass[final]{cvpr}

\usepackage{times}
\usepackage{epsfig}
\usepackage{graphicx}
\usepackage{amsmath}
\usepackage{amssymb}

\usepackage{multirow}
\usepackage{extarrows}
\usepackage{framed}
\usepackage{booktabs}
\usepackage{gensymb} 
\usepackage{url}
\usepackage{threeparttable}
\usepackage{tablefootnote}
\usepackage{color}

\usepackage[pagebackref=true,breaklinks=true,colorlinks,bookmarks=false]{hyperref}

\def\cvprPaperID{7407} 
\def\confYear{CVPR 2021}
\begin{document}

\title{Verifiability and Predictability: Interpreting Utilities of Network Architectures \\for Point Cloud Processing}

\author{Wen Shen$^{2,*}$,
	Zhihua Wei$^{2,*}$,
	Shikun Huang$^{2}$,
	Binbin Zhang$^{2}$,
	Panyue Chen$^{2}$,
	Ping Zhao$^{2}$, \\
	Quanshi Zhang$^{1,\dag}$ \\
	$^1$Shanghai Jiao Tong University, Shanghai, China \\
	$^2$Tongji University, Shanghai, China \\
	{\tt\small \{wen\_shen,zhihua\_wei,hsk,0206zbb,2030793,zhaoping\}@tongji.edu.cn,zqs1022@sjtu.edu.cn}
}	


\maketitle

\pagestyle{empty}  
\thispagestyle{empty} 

{
	\renewcommand{\thefootnote}{\fnsymbol{footnote}}
\footnotetext[1]{Wen Shen and Zhihua Wei have equal contributions.}
\footnotetext[2]{Quanshi Zhang is the corresponding author. This work is conducted under the supervision of Dr. Quanshi Zhang. He is with the John Hopcroft Center and the MoE Key Lab of Artificial Intelligence, AI Institute, at the Shanghai Jiao Tong University, China.}
}

\begin{abstract}
   In this paper, we diagnose deep neural networks for 3D point cloud processing to explore utilities of different intermediate-layer network architectures. We propose a number of hypotheses on the effects of specific intermediate-layer network architectures on the representation capacity of DNNs. In order to prove the hypotheses, we design five metrics to diagnose various types of DNNs from the following perspectives, information discarding, information concentration, rotation robustness, adversarial robustness, and neighborhood inconsistency. We conduct comparative studies based on such metrics to verify the hypotheses. We further use the verified hypotheses to revise intermediate-layer architectures of existing DNNs and improve their utilities. Experiments demonstrate the effectiveness of our method. \emph{The code will be released when this paper is accepted.}
\end{abstract}

\section{Introduction}
Recently, a series of works use deep neural networks (DNNs) for 3D point cloud processing and have achieved superior performance in various 3D tasks. However, traditional studies usually designed intermediate-layer architectures based on empiricism. Exploring and verifying utilities of each specific intermediate-layer architecture from the perspective of a DNN's representation capacity still present significant challenges for state-of-the-art algorithms.

In this study, we aim to bridge the gap between the intermediate-layer architecture and its utilities. Table~\ref{t1} lists three kinds of utilities considered in this study, including rotation robustness, adversarial robustness, and neighborhood inconsistency. Although there are many heuristic insights on utilities of existing architectures for 3D point cloud processing, there does not exist a rigorous and quantitative verification of such insights.

Therefore, we propose a method to quantitatively diagnose the utilities of intermediate-layer network architectures, which will provide new insights into architectural design. This is a necessary step towards the deep learning with scientific rigour. Note that, utilities are not necessarily equivariant to advantages. For example, in most cases, the rotation robustness is supposed to be a good property. However, the rotation robustness sometimes requires a DNN not to encode rotation-sensitive but discriminative features.

\begin{table}[t]
	\begin{center}
		\includegraphics[width=0.65\linewidth]{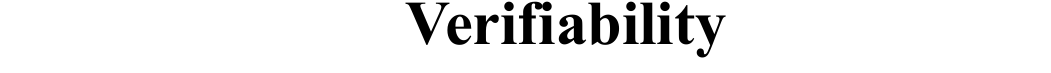}
		\newcommand{\tabincell}[2]{\begin{tabular}{@{}#1@{}}#2\end{tabular}}
		\resizebox{\linewidth}{!}{\begin{tabular}{l|ccc}
				\hline
				\multicolumn{1}{c|}{Intermediate-layer architectures} &\!\!\!\! \tabincell{c}{Rotation \\ robustness}  \!\!\!\!&\!\!\!\! \tabincell{c}{Adversarial \\ robustness} \!\!\!\!&\!\!\!\! \tabincell{c}{Neighborhood \\consistency} \\
				\hline
				\tabincell{l}{(a) Modules of using information of \\ $\quad\ $ local density to reweight features \cite{wu2019pointconv}.}  \!\!\!\!&\!\!\!\! --  \!\!\!\!&\!\!\!\! \checkmark \!\!\!\!&\!\!\!\! --  \\
				\tabincell{l}{(b) Modules of using information of \\ $\quad\ $ local coordinates to reweight features \cite{wu2019pointconv}.} \!\!\!\!&\!\!\!\! \checkmark \!\!\!\!&\!\!\!\!-- \!\!\!\!&\!\!\!\! -- \\
				\tabincell{l}{	(c) Modules of concatenating \\ $\quad\ $ multi-scale features \cite{liu2018point2sequence}.}  \!\!\!\!&\!\!\!\! -- \!\!\!\!&\!\!\!\! \checkmark \!\!\!\!&\!\!\!\! \checkmark \\
				\tabincell{l}{(d) Modules of computing \\$\quad\ $ orientation-aware features \cite{jiang2018pointsift}.} \!\!\!\!&\!\!\!\! \checkmark  \!\!\!\!&\!\!\!\! -- \!\!\!\!&\!\!\!\!--  \\
				\hline
		\end{tabular}}
		\includegraphics[width=0.65\linewidth]{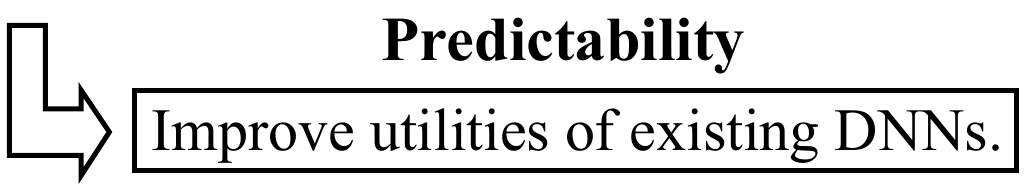}
	\end{center}
	\caption{Illustration of the verified utilities of specific intermediate-layer architectures. ``--'' denotes that the utility has not been examined, instead of indicating non-existence of the utility. Please see Fig.~\ref{fig:architectures} for architectural details.}
	\label{t1}
\end{table}

This study focuses on two terms, \emph{i.e.} verifiability and predictability. In terms of verifiability, we design new metrics to quantify utilities of existing intermediate-layer architectures to prove intuitive insights. In terms of predictability, we further use the verified insights to revise other networks to improve their utilities. Note that, the revision of intermediate-layer architectures generally dose not change the depth of DNNs, so that we eliminate the influence of the depth change.

More specifically, we propose a few hypotheses of utilities of specific intermediate-layer architectures, as shown in Table~\ref{t1}. Theoretically, we could analyze specific intermediate-layer architectures \emph{w.r.t.} all utilities. However, due to the limit of the page number, we only verify hypotheses with stong connection to human intuitions. We design and conduct comparative studies to verify these hypotheses. The verified hypotheses are further used to guide the architectural revision of existing DNNs to improve their utilities. The verified hypotheses can be summarized as follows.

$\bullet\;$The specific module in \cite{wu2019pointconv}, which uses the local density information to reweight features (Fig.~\ref{fig:architectures} (a)), improves the adversarial robustness (Table~\ref{t1} (a)).

$\bullet\;$Another specific module in \cite{wu2019pointconv}, which uses local 3D coordinates' information to reweight features (Fig.~\ref{fig:architectures} (b)), improves the rotation robustness (Table~\ref{t1} (b)).

$\bullet\;$The specific module in \cite{qi2017pointnet++, liu2018point2sequence}, which extracts multi-scale features (Fig.~\ref{fig:architectures} (c)), improves the adversarial robustness and the neighborhood consistency (Table~\ref{t1} (c)). Neighborhood consistency measures whether a DNN assigns similar attention to neighboring points.

$\bullet\;$The specific module in \cite{jiang2018pointsift}, which encodes the information of different orientations (Fig.~\ref{fig:architectures} (d)), improves the rotation robustness (Table~\ref{t1} (d)).

In order to verify the above hypotheses, we design the following five evaluation metrics and conduct a number of comparative experiments to quantify utilities of different intermediate-layer architectures.

\textit{1. Information discarding and 2. information concentration:}
Information discarding measures how much information of an input point cloud is forgotten during the computation of a specific intermediate-layer feature. From the perspective of information propagation, the forward propagation through layers can be regarded as a hierarchical process of discarding input information \cite{shwartz2017opening}. Ideally, a DNN is supposed to discard information that is not related to the task. Let us take the task of object classification for example. The information of foreground points is usually supposed to be related to the task, while that of background points is not related to the task and is discarded.

To this end, we further propose information concentration to measure the gap between the information related to the task and the information not related to the task. Information concentration can be used to evaluate a DNN's ability to focus on points related to the task.

\textit{3. Rotation robustness:}
Rotation robustness measures whether a DNN will use the same logic to recognize the same object when a point cloud has been rotated by a random angle. In other words, if two point clouds have the same global shape but different orientations, the DNN is supposed to select the same regions/points to compute the intermediate-layer feature. Unlike images with rich color information, point clouds usually only use spatial contexts for classification. Therefore, a well-trained DNN is supposed to have the rotation robustness.

\textit{4. $\ \ $Adversarial robustness:}
A reliable DNN is supposed to be robust to adversarial attacks.

\textit{5. Neighborhood inconsistency\footnote{Values of the rotation robustness and the neighborhood consistency are negative numbers. For intuitive comparisons, we showed results of rotation non-robustness and neighborhood inconsistency in Section~\ref{sec:exp}.}:}
Neighborhood inconsistency measures whether adjacent points have similar importance in the computation of an intermediate-layer feature. Adjacent points in a 3D object usually have similar shape contexts, so they are supposed to have similar importance. Therefore, ideally, a well-trained DNN should have a low value of neighborhood inconsistency.

The verified hypotheses are then applied to existing DNNs to revise their intermediate-layer architectures and improve their utilities. Note that this study aims to verify some insights about intermediate-layer architectures in the scenario of object classification, in order to improve utilities of existing DNNs. The classification accuracy is reported in supplementary materials.

Note that in comparative studies, unnecessarily complex intermediate-layer architectures usually bring in additional uncertainty, which will prevent our experiments from obtaining reliable and rigorous results. Therefore, we conduct experiments on simple-yet-classic intermediate-layer architectures.

Contributions of our study are summarized as follows. (1) We propose a few hypotheses on utilities of specific intermediate-layer architectures. (2) We design five metrics to conduct comparative studies to verify these hypotheses, which provide new insights into architectural utilities. (3) It is proved that the verified hypotheses can be used to revise existing DNNs to improve their utilities.

\section{Related work}

\textbf{Deep learning on 3D Point Cloud:} Recently, many approaches use DNNs for 3D point cloud processing and have exhibited superior performance in various 3D tasks \cite{qi2017pointnet,su2018splatnet,valsesia2018learning,yu2018pu,yang2018foldingnet,gadelha2018multiresolution,wang2018sgpn,Komarichev_2019_CVPR,shi2019pointrcnn,liu2019deep,shen20193d}. PointNet \cite{qi2017pointnet} was a pioneer in this direction, which used a max pooling layer to aggregate all individual point features into a global feature. However, such architecture fell short of capturing local features. PointNet++ \cite{qi2017pointnet++} hierarchically used PointNet as a local descriptor to extract contextual information. Some studies \cite{jiang2018pointsift,wang2018dynamic,Komarichev_2019_CVPR,liu2019relation} further improved the networks' ability to capture local geometric features. Some studies used graph convolutional neural networks for 3D point cloud processing \cite{simonovsky2017dynamic,wang2018dynamic}. Others focused on the correlations between different regions of the 3D point cloud \cite{liu2018point2sequence} or interaction between points \cite{Zhao_2019_CVPR}. \
In comparison, our study focuses on the utility analysis of intermediate-layer network architectures for point cloud processing.

\textbf{Visualization or diagnosis of representations:} The most intuitive way to interpret DNNs is the visualization of visual patterns corresponding to a feature map or the network output \cite{zeiler2014visualizing,mahendran2015understanding,dosovitskiy2016inverting,zhou2014object}, such as gradient-based methods \cite{fong2018net2vec,selvaraju2017grad}, and the estimation of the saliency map \cite{ribeiro2016should,lundberg2017unified,kindermans2017learning,qi2017pointnet,zheng2019pointcloud}. In comparison, our study aims to explore the utility of intermediate-layer network architectures by diagnosing the information-processing logic of DNNs.

\textbf{Quantitative evaluation of representations:} In the field of explainable AI, explaining the capasity of representations has attracted increasing research attention.
Some studies aimed to disentangle features of a DNN into quantifiable and interpretable feature components \cite{zhang2019interpreting,zhang2018interpretable,zhang2018interpreting}. Some studies quantified the representation similarity to help understand the neural networks \cite{gotmare2018closer,kornblith2019similarity,morcos2018insights,raghu2017svcca}. Li \etal \cite{li2019exploiting} quantitated the importance of different feature dimensions to guide model compression. Zhang \etal \cite{zhang2020interpreting} quantitated the significance of interactions among multiple input variables of the DNN. Other studies explained the representation capacity of DNNs \cite{cheng2020explaining,liang2019knowledge,wang2020unified,zhang2020interpreting}. The information-bottleneck theory \cite{tishby2000information,shwartz2017opening,cheng2018evaluating} explained the trade-off between the information compression and the discrimination power of features in a neural network. Achille and Soatto \cite{achille2018information} designed an information Dropout layer and quantified the information transmitted through it. Ma \etal \cite{ma2019quantifying} presented a method to calculate the entropy of the input information. Inspired by \cite{ma2019quantifying}, we propose five metrics to diagnose feature representations of different DNNs and explore utilities of different intermediate-layer network architectures.

\section{Metrics to Diagnose Networks}

\subsection{Preliminaries: quantification of entropy-based information discarding}\label{sec:3-1}
We extend the method of calculating the entropy of the input information, which is proposed in \cite{ma2019quantifying}, as the technical foundation. Based on this, a number of new metrics are designed to diagnose the DNN. The method quantifies the discarding of the input information during the layerwise forward propagation by computing the entropy of the input information given the specific feature of an intermediate layer. Given a point cloud {$X$}, let {\small$f=h(X)$} denote the feature of a specific intermediate layer. It is assumed that {\small$f'$} represents the same object concept\footnote{In this study, the concept of an object is referred to as a small range of features that represent the same object instance.} as {\small$f$} when {\small$f'$} satisfies {\small$\left\|f'-f\right\|^2 < \epsilon$}, where feature {\small$f'=h({X'})$}, {\small$X'=X+\boldsymbol{\delta}$}. {\small$\boldsymbol{\delta}$} denotes a random noise. Given a specific feature, the conditional entropy of the input information is computed, when the input represents a specific object concept. \emph{I.e.} we calculate entropy {\small$H(X')$}, \emph{s.t.} {\small$\left\|f'-f\right\|^2 < \epsilon$}. It is assumed that {\small$X'$} follows a Gaussian distribution {\small$X'\sim\mathcal{N}(X,\Sigma=diag[\sigma_1^2,\sigma_2^2,\dots])$}. {\small$\Sigma$} measures the maximum perturbation added to {\small$X$} following the maximum-entropy principle, which subjects to {\small$\left\|f'-f\right\|^2 < \epsilon$}. Considering the assumption of the i.i.d. dimensions of {\small$X'$}, the overall entropy {\small$H(X')$} can be decomposed into point-wise entropies.
	\begin{equation}\label{sid}
	\max\limits_{\boldsymbol{\sigma} = [\sigma_1,\sigma_2,\dots]^{\top}} H(X'), \quad {\rm s.t.} \ \left\|h(X')-f\right\|^2 < \epsilon\footnote{We follow the \cite{ma2019quantifying} to slightly adjust the value of {\small$\lambda$} to make the learned {\small$\boldsymbol{\sigma}$} satisfy that  {\small$\mathbb{E}_{f'}[\left\|f'-f\right\|^2]$} is about twice the inherent variance of intermediate-layer features subject to a small input noise. {\small$\lambda$} is a hyperparameter defined in \cite{ma2019quantifying}.},\qquad  \\
	\end{equation}
where {\small$H(X')=\sum_{i}H_i$}; {\small$H_i=\log\sigma_i+\tfrac{1}{2}\log(2\pi e)$} denotes the entropy of the {\small$i$}-th point. {\small$H_i$} quantifies how much information of the {\small$i$}-th point can be discarded, when the feature {\small$h(X')$} is required to represent the concept of the target object.

\subsection{Five metrics}\label{sec:fivemetrics}

\textbf{Metric 1, information discarding:} The information discarding is defined as {\small$H(X')$} in Eqn.~(\ref{sid}). The information discarding is measured at the point level, \emph{i.e.} {\small$H_i$}, which quantifies how much information of the {\small$i$}-th point is discarded during the computation of an intermediate-layer feature. The point with a lower value of {\small$H_i$} is regarded more important in the computation of the feature.

\textbf{Metric 2, information concentration:} The information concentration is based on the metric of information discarding. The information concentration is used to analyze a DNN's ability to maintain the input information related to the task, and discard redundant information unrelated to the task. \emph{E.g.}, in the task of object classification, background points are usually supposed not to be related to the task and are therefore more likely to be discarded by the DNN. Let {\small$\Lambda^{\rm foreground}$} denote the set of points in the foreground object in the point cloud {\small$X$}, and let {\small$\Lambda^{\rm background}$} denote the set of points in the background. Information concentration can be computed as the relative background information discarding \emph{w.r.t.} foreground information discarding.
	\begin{equation}
	\label{cic}
	\mathbb{E}_{i\in\Lambda^{\rm background}}[H_i]-\mathbb{E}_{i\in\Lambda^{\rm foreground}}[H_i],
	\end{equation}
where a higher value of information concentration indicates that the DNN concentrates more on the foreground points during the computation of the feature.

Note that most widely used benchmark datasets for point cloud classification only contain foreground objects. Therefore, we generate a new dataset, where each point cloud contains both the foreground object and the background. In this new dataset, the background is composed of points that are irrelevant to the foreground. We will introduce details in Section~\ref{sec:exp}.

\begin{figure*}[tbp]
	\begin{center}
		\includegraphics[width=\linewidth]{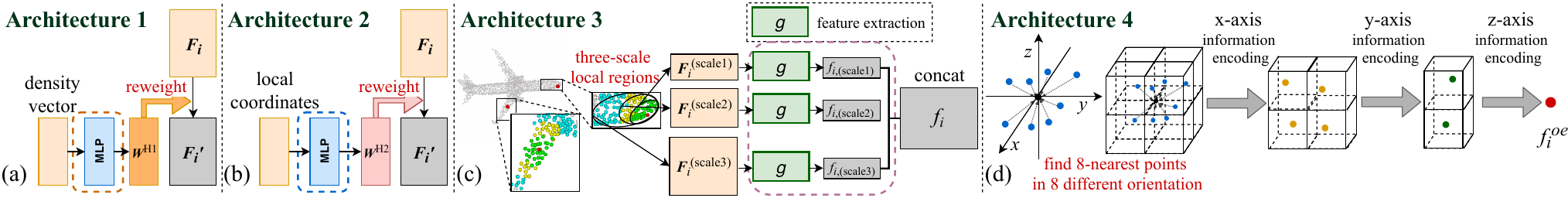}
	\end{center}
	\caption{Illustration of the specific intermediate-layer architectures. Please see texts in Section~\ref{sec:intro-architectures} for architectural details.}
	\label{fig:architectures}
\end{figure*}

\textbf{Metric 3, rotation robustness:}
Given the two point clouds with the same shape but different orientations, the rotation robustness is proposed to measure whether a DNN uses similar subsets of two point clouds to compute the intermediate-layer feature. Let {\small$X_1=\theta_1(X)$} and {\small$X_2=\theta_2(X)$} denote the point clouds that have the same shape but different orientations, where {\small$\theta_1$} and {\small$\theta_2$} denote two different rotation operations. To quantify the similarity of the attention on the two point clouds, we compute the Jensen-Shannon divergence between the distributions of information discarding in Eqn.~(\ref{sid}), {\small$\boldsymbol{\delta}_1$} and {\small$\boldsymbol{\delta}_2$} \emph{w.r.t.} {\small$X'_1 =\theta_1(X+\boldsymbol{\delta}_1)$}, {\small$X'_2 =\theta_2(X+\boldsymbol{\delta}_2)$}. \emph{I.e.} we measure whether the DNN ignores similar sets of points to compute features of the two point clouds.
	\begin{equation}\label{rr}
	JSD(\boldsymbol{\delta}_1||\boldsymbol{\delta}_2),\quad
	\textrm{s.t.} \quad
	\begin{cases}
	 ||h(X'_1)-h(X_1)||^2<\epsilon\\
	  ||h(X'_2)-h(X_2)||^2<\epsilon
	 \end{cases}
	\end{equation}
where {\small$JSD(\boldsymbol{\delta}_1||\boldsymbol{\delta}_2)$} measures the dissimilarity between information distributions over the two point clouds.

The rotation non-robustness\textcolor{red}{\footnotemark[1]} is defined as the average of the dissimilarity of attention on any two point clouds with different orientations, \emph{i.e.} {\small$\mathbb{E}_{\forall {\theta_1},{\theta_2}}[JSD(\boldsymbol{\delta}_1||\boldsymbol{\delta}_2)]$}. In this study, we use the variational-approximation-based method in \cite{hershey2007approximating} to approximate the Jensen-Shannon divergence.

\textbf{Metric 4, adversarial robustness:}
We use the method in \cite{szegedy2013intriguing} to perform adversarial attacks. The objective is
	\begin{equation}
	\min \lVert\boldsymbol{\epsilon}\lVert_2^2, \qquad \textrm{s.t.} \quad C(X+ \boldsymbol{\epsilon}  ) = \hat{l} \ne l^*,
	\end{equation}
where {\small$C(\cdot)$} is the predicted label; {\small$l^*$} is the correct label of {\small$X$}; {\small$\hat{l}$} is a target incorrect label. In this study, we perform targeted adversarial attacks against all incorrect classes. We use the average of {\small$\lVert\boldsymbol{\epsilon}\lVert_2$} over all incorrect classes to measure the adversarial robustness.

\textbf{Metric 5, neighborhood inconsistency\textcolor{red}{\footnotemark[1]}}:
The neighborhood inconsistency is proposed to evaluate a DNN's ability to assign similar attention to neighboring points during the computation of an intermediate-layer feature. Ideally, for a DNN, except for special points (\emph{e.g.} those on the edge), most neighboring points in a small region of a point cloud usually have similar shape contexts, so they are supposed to make similar contributions to the classification and receive similar attention, \emph{i.e.} low neighborhood inconsistency. Let {\small$\mathbf{N}(i)$} denote a set of {\small$K$} nearest points of the {\small$i$}-th point. We define the neighborhood inconsistency as the difference between the maximum and minimum point-wise information discarding within {\small$\mathbf{N}(i)$}.
	\begin{equation}
	\mathbb{E}_{i}[{\rm max}_{j\in\mathbf{N}(i)}H_j-{\rm min}_{j\in\mathbf{N}(i)}H_j].
	\end{equation}

\section{Hypotheses and Comparative Study}
\subsection{Overview of intermediate-layer architectures}\label{sec:intro-architectures}

$\bullet\;$\textbf{Notation:} Let {\small$x_i\in \mathbb{R}^{3}$} denote the {\small$i$}-th point, {\small$i=1,2,\dots,n$}; let {\small$\mathbf{N}(i)$} denote a set of {\small$K$} nearest points of {\small$x_i$}; let {\small$\mathbf{F}_i\in \mathbb{R}^{d\times K}$} denote intermediate-layer features of neighboring points {\small$\mathbf{N}(i)$}, where each column of {\small$\mathbf{F}_i$} represents the feature of a specific point in {\small$\mathbf{N}(i)$}.

$\bullet\;$\textbf{Architecture 1, features reweighted by the information of the local density:} Architecture 1 focuses on the use of the local density information to reweight features \cite{wu2019pointconv}. As shown in Fig.~\ref{fig:architectures} (a), for each point {\small$x_i$}, Architecture 1 uses the local density \emph{w.r.t.} neighboring points of {\small$x_i$} to compute {\small$\boldsymbol{W}^{\rm{H_1}}\in \mathbb{R}^{K}$}, which reweights intermediate-layer features {\small$\mathbf{F}_i$}.
	\begin{equation}\label{ids}
	\mathbf{F}'_i= \mathbf{F}_i\ {diag}[\boldsymbol{W}^{\rm{H_1}}], \ \ \    \boldsymbol{W}^{\rm{H_1}}\!\!=mlp(density(\mathbf{N}(i))),
	\end{equation}
where {\small$diag[\boldsymbol{W}^{\rm H_1}]$} transforms the vector {\small$\boldsymbol{W}^{\rm H_1}$} into a diagonal matrix; {\small$density(\mathbf{N}(i))$} is a vector representing the density of neighboring points in {\small$\mathbf{N}(i)$}; {\small$mlp$} is a two-layer perceptron network.

$\bullet\;$\textbf{Architecture 2, features reweighted by the information of local coordinates:} As shown in Fig.~\ref{fig:architectures} (b), for each point {\small$x_i$}, Architecture 2 uses the information of local 3D coordinates to compute {\small$\boldsymbol{W}^{\rm{H_2}}\in \mathbb{R}^{M\times K}$} to reweight intermediate-layer features {\small${\bf F}_i$}.
	\begin{equation}\label{3dw}
	{\bf F}'_i={\bf F}_i(\boldsymbol{W}^{\rm{H_2}})^{\top}, \quad \boldsymbol{W}^{\rm{H_2}}=mlp(\{x_j|j\in\mathbf{N}(i)\}),
	\end{equation}
where the {\small$mlp$} is a single-layer perceptron network.

$\bullet\;$\textbf{Architecture 3, multi-scale features:} Architecture 3 focuses on the use of multi-scale contextual information \cite{qi2017pointnet++,liu2018point2sequence}. As illustrated in Fig.~\ref{fig:architectures} (c), {\small$\{{\bf F}^{{\rm scale}=K_1}_i, ..., {\bf F}^{{\rm scale}=K_T}_i\}$} denote features that are extracted using contexts of {\small$x_i$} at different scales, {\small${\bf F}^{{\rm scale}=K_t}_i\in\mathbb{R}^{d\times K_t}$}. Each specific context \emph{w.r.t.} {\small$x_i$} is composed of {\small$K_t$} nearest neighboring points around {\small$x_i$}. Then, {\small$f_{i,({\rm scale}=K_t)}^{\rm upper}\in \mathbb{R}^D$} in the upper layer is computed using {\small${\bf F}_i^{{\rm scale}=K_t}$}. Architecture 3 concatenates these multi-scale features to obtain {\small$f_i^{\rm upper}$}.
	\begin{equation}\label{ms}
	\begin{aligned}
	&f_i^{\rm upper}=concat
	\begin{Bmatrix}
	f_{i,({\rm scale}=K_1)}^{\rm upper}\\
	f_{i,({\rm scale}=K_2)}^{\rm upper} \\
	\cdots \\
	f_{i,({\rm scale}=K_T)}^{\rm upper} \\
	\end{Bmatrix},\\
	&\quad f_{i,({\rm scale}=K_t)}^{\rm upper}=g({\bf F}_i^{{\rm scale}=K_t}),
	\end{aligned}
	\end{equation}
where {\small$g(\cdot)$} is a function for feature extraction. Details about this function are introduced in \cite{qi2017pointnet}, which are also summarized in supplementary materials\footnote{For the convenience of readers to quickly understand relevant technologies in original papers, we summarize these relevant technologies in supplementary materials.}.

$\bullet\;$\textbf{Architecture 4, orientation-aware features:} Architecture 4 focuses on the use of orientation information \cite{jiang2018pointsift}. As illustrated in Fig.~\ref{fig:architectures} (d), for each point {\small$x_i$}, {\small${\bf F}_i^{\rm oe}\in \mathbb{R}^{d\times O}$} denotes the feature of {\small$x_i$}, which encodes the information of various orientations, where {\small$O$} is the number of orientations. Architecture 4 uses {\small${\bf F}_i^{\rm oe}$} to compute the orientation-aware feature {\small$f_i^{\rm oe}\in \mathbb{R}^d$}.
	\begin{equation}\label{oe}
	f_i^{\rm oe}=Conv^{\rm oe}({\bf F}_i^{\rm oe}),
	\end{equation}
where {\small$Conv^{\rm oe}$} is a special convolution operator. Details about this operator and the computation of {\small$f_i^{\rm oe}$} are introduced in \cite{jiang2018pointsift}\textcolor{red}{\footnotemark[3]}.

\subsection{Four hypotheses and comparative study design}\label{sec:hypotheses}

\begin{framed}
\vspace{-2pt}
\textbf{Hypothesis 1:} Architecture 1 designed by \cite{wu2019pointconv} (in Fig.~\ref{fig:architectures} (a)) increases adversarial robustness.\vspace{-2pt}
\end{framed}

This hypothesis is based on the observation that PointConv \cite{wu2019pointconv} has strong adversarial robustness, which may stem from Architecture 1. To verify this, we construct two versions of the PointConv for comparison, \emph{i.e.} one with Architecture 1 and the other without Architecture 1.

To obtain the PointConv without Architecture 1, we remove\footnote{Note that removing or adding modules of these specific intermediate-layer network architectures generally has no effects on the depth of DNNs, so that we eliminate the influence of changes in DNNs' depth.} all the modules of Architecture 1 from the original network (see the footnote\footnote{The PointConv for classification is revised from the code for segmentation released by \cite{wu2019pointconv}.}), which are located behind the 2-nd, 5-th, 8-th, 11-th, and 14-th nonlinear transformation layers. The global architecture of PointConv is introduced in \cite{wu2019pointconv}\textcolor{red}{\footnotemark[3]}.
	\begin{equation}\label{eq:conv1}
	\begin{aligned}
	f_i^{\rm upper}\!\!= mlp({\bf F}_i)\ diag[\boldsymbol{W}^{\rm H_1}] \ \Longrightarrow \
	f_i^{\rm upper}\!\!=mlp({\bf F}_i),
	\end{aligned}
	\end{equation}
where {\small$f_i^{\rm upper}$} is the feature in the upper layer; {\small$diag[\boldsymbol{W}^{\rm H_1}]$} transforms the vector {\small$\boldsymbol{W}^{\rm H_1}$} into a diagonal matrix.

\begin{framed}
\vspace{-2pt}
\textbf{Hypothesis 2:} Architecture 2 designed by \cite{wu2019pointconv} (in Fig.~\ref{fig:architectures} (b)) increases rotation robustness.\vspace{-2pt}
\end{framed}

This hypothesis is proposed based on the observation that PointConv \cite{wu2019pointconv} has strong rotation robustness, which may stem from Architecture 2. To verify this, we construct two versions PointConv for comparison, \emph{i.e.} one with Architecture 2 and the other without Architecture 2.

To obtain the PointConv without Architecture 2, we remove\textcolor{red}{\footnotemark[4]} all the modules of Architecture 2, which are located before the 3-rd, 6-th, 9-th, 12-th, and 15-th nonlinear transformation layers. The global architecture of PointConv is introduced in \cite{wu2019pointconv}\textcolor{red}{\footnotemark[3]}.
	\begin{equation}\label{eq:conv2}
	\begin{aligned}
	f_i^{\rm upper}=mlp({\bf F}_i)(\boldsymbol{W}^{\rm H_2})^{\top} \  \Longrightarrow \ f_i^{\rm upper}=mlp({\bf F}_i).
	\end{aligned}
	\end{equation}

\begin{framed}
\vspace{-2pt}
\textbf{Hypothesis 3:} Architecture 3 used in \cite{qi2017pointnet++,liu2018point2sequence} (in Fig.~\ref{fig:architectures} (c)) increases adversarial robustness and neighborhood consistency.\vspace{-2pt}
\end{framed}

This hypothesis is inspired by \cite{qi2017pointnet++, liu2018point2sequence}, which encodes multi-scale contextual information. To verify this hypothesis, we construct\textcolor{red}{\footnotemark[4]} three versions of Point2Sequence for comparison. The baseline network of Point2Sequence concatenates features of 4 different scales to compute the feature in the upper layer, {\small$\{f_{i,({\rm scale}=K_1)}^{\rm upper},f_{i,({\rm scale}=K_2)}^{\rm upper},f_{i,({\rm scale}=K_3)}^{\rm upper},f_{i,({\rm scale}=K_4)}^{\rm upper}\}$}. In this study, we set {\small$K_1$} = 128, {\small$K_2$} = 64, {\small$K_3$} = 32, and {\small$K_4$} = 16. The first network extracts features with three different scales, {\small\{$f_{i,({\rm scale}=K_1)}^{\rm upper},f_{i,({\rm scale}=K_2)}^{\rm upper},f_{i,({\rm scale}=K_3)}^{\rm upper}$\}}, and the second one extracts features with two different scales, {\small\{$f_{i,({\rm scale}=K_1)}^{\rm upper},f_{i,({\rm scale}=K_2)}^{\rm upper}$\}}. The global architecture of Point2Sequence is introduced in \cite{liu2018point2sequence}\textcolor{red}{\footnotemark[3]}.

\begin{framed}
\vspace{-2pt}
\textbf{Hypothesis 4:} Architecture 4 designed by \cite{jiang2018pointsift} (in Fig.~\ref{fig:architectures} (d)) increases rotation robustness.\vspace{-2pt}
\end{framed}

This hypothesis is proposed based on the observation that PointSIFT \cite{jiang2018pointsift} exhibits strong rotation robustness. It may be because Architecture 4 ensures that features contain information from various orientations. To verify this hypothesis, we construct two versions of the PointSIFT for comparisons, \emph{i.e.} one with Architecture 4 and the other without Architecture 4.

To get the PointSIFT without Architecture 4, we remove all modules of Architecture 4 from the original network (see the footnote\footnote{The PointSIFT for classification is revised from the code for segmentation released by \cite{jiang2018pointsift}.}), which are located before the 1-st, 3-rd, 5-th, and 7-th nonlinear transformation layers. The global architecture of PointSIFT is introduced in \cite{jiang2018pointsift}\textcolor{red}{\footnotemark[3]}.

\begin{table*}[t]
	\begin{center}
		\resizebox{\linewidth}{!}{\begin{threeparttable}
				\begin{tabular}{cl|c|c}
					\toprule[1pt]
					& &\# of added modules&Locations of added modules\\
					\hline
					\multirow{3}{*}{(a)} &Add Architecture 1 to PointNet++ for adversarial robustness
					& 3 & behind the 3-rd, 6-th, and 9-th nonlinear transformation layers \\
					\cline{2-4}
					&Add Architecture 1 to Point2Sequence for adversarial robustness
					&1&behind the last nonlinear transformation layer\\
					\cline{2-4}
					&Add Architecture 1 to RSCNN for adversarial robustness
					&2 & behind the 2-nd and 5-th nonlinear transformation layers\\
					\hline
					\multirow{3}{*}{(b)}&Add Architecture 2 to PointNet++ for rotation robustness
					&3&behind the 3-rd, 6-th, and 9-th nonlinear transformation layers\\
					\cline{2-4}
					&Add Architecture 2 to Point2Sequence for rotation robustness
					&1&behind the last nonlinear transformation layer\\
					\cline{2-4}
					&Add Architecture 2 to RSCNN for rotation robustness
					&2& behind the 2-nd and 5-th nonlinear transformation layers\\
					\hline
					
					\multirow{4}{*}{(c)}&Add Architecture 3 to PointNet++ for adversarial robustness
					&\multirow{2}{*}{2}& for 1-st to 3-rd nonlinear transformation layers, {\small$K_1$}=16, {\small$K_2$} =128\\
					&\& neighborhood consistency
					& & for 4-th to 6-th nonlinear transformation layers, {\small$K_1$}=32, {\small$K_2$} =128\tnote{*}\\
					
					\cline{2-4}
					&Add Architecture 3 to RSCNN for adversarial robustness
					&\multirow{2}{*}{2}& for 1-st to 3-rd nonlinear transformation layers, {\small$K_1$}=16, {\small$K_2$} =32\\
					&\& neighborhood consistency
					& & for 4-th to 6-th nonlinear transformation layers, {\small$K_1$}=16, {\small$K_2$} =48\tnote{*}\\
					
					\hline
					\multirow{3}{*}{(d)}&Add Architecture 4 to PointNet++ for rotation robustness
					&1&before the 7-th nonlinear transformation layer\\
					\cline{2-4}
					&Add Architecture 4 to Point2Sequence for rotation robustness
					&1&before the 14-th nonlinear transformation layer \\
					\cline{2-4}
					&Add Architecture 4 to RSCNN for rotation robustness
					&1&before the 7-th nonlinear transformation layer \\
					\bottomrule[1pt]
				\end{tabular}	
				\begin{tablenotes}
					\item[*] {\small$K_1$} and {\small$K_2$} are hyper-parameters of added modules of Architecture 3, which have been introduced in Eqn.~(\ref{ms}).
				\end{tablenotes}
		\end{threeparttable} }
	\end{center}
\caption{Adding\textcolor{red}{\protect\footnotemark[4]} specific intermediate-layer architectures to existing DNNs to improve utilities.}
\label{tab:comparison}
\end{table*}

\subsection{Comparative study for the improvement of utilities of existing DNNs}\label{sec:utility}

In this section, we further prove that the verified four hypotheses can be used to revise existing intermediate-layer network architectures in order to improve their utilities. We apply our method to three benchmark DNNs, including PointNet++ \cite{qi2017pointnet++}, Point2Sequence \cite{liu2018point2sequence}, and RSCNN \cite{liu2019relation}.

In Section~\ref{sec:hypotheses}, we remove specific intermediate-layer architectures from original DNNs. Actually, if we take the DNN without the specific intermediate-layer architecture as the original one (\emph{e.g.} the PointConv without Architecture 1) and take the real original DNN as the revised one (\emph{e.g.} the PointConv with Architecture 1), then it is naturally proved that the verified hypotheses can be used to revise DNNs to improve their utilities.

Nevertheless, in this section, we further prove that these specific intermediate-layer architectures can improve utilities of other DNNs.

\begin{framed}
\vspace{-2pt}
\textbf{Architecture 1} designed by \cite{wu2019pointconv} is added to PointNet++, Point2Sequence, and RSCNN in order to improve their adversarial robustness.\vspace{-2pt}
\end{framed}

For each network, we construct two versions for comparison, \emph{i.e.} one with Architecture 1 and the other without Architecture 1.

Table \ref{tab:comparison} (a) shows details about how to obtain DNNs with Architecture 1. Global architectures of PointNet++, Point2Sequence, and RSCNN are introduced in \cite{qi2017pointnet++}\textcolor{red}{\footnotemark[3]}, \cite{liu2018point2sequence}\textcolor{red}{\footnotemark[3]}, and \cite{liu2019relation}\textcolor{red}{\footnotemark[3]}.

Eqn. (\ref{eq:plus1}) shows how to add Architecture 1 behind a nonlinear transformation layer.
	\begin{equation}\label{eq:plus1}
	\begin{aligned}
	f_i^{\rm upper}\!\!=mlp({\bf F}_i)\ \Longrightarrow \
	f_i^{\rm upper}\!\!=mlp({\bf F}_i)\ diag[\boldsymbol{W}^{\rm H_1}],
	\end{aligned}
	\end{equation}
where {\small$\boldsymbol{W}^{\rm H_1}$} denotes the formula of Architecture 1, which has been introduced in Eqn.~(\ref{ids}).

\begin{framed}
\vspace{-2pt}
\textbf{Architecture 2} designed by \cite{wu2019pointconv} is added to PointNet++, Point2Sequence, and RSCNN in order to improve their rotation robustness.\vspace{-2pt}
\end{framed}
\vspace{-4pt}

Table \ref{tab:comparison} (b) shows details about how to obtain DNNs with Architecture 2. Global architectures of PointNet++, Point2Sequence, and RSCNN are introduced in \cite{qi2017pointnet++}\textcolor{red}{\footnotemark[3]}, \cite{liu2018point2sequence}\textcolor{red}{\footnotemark[3]}, and \cite{liu2019relation}\textcolor{red}{\footnotemark[3]}.

Eqn. (\ref{flocal-wh2}) shows how to add Architecture 2 behind a nonlinear transformation layer.
	\begin{equation}\label{flocal-wh2}
	\begin{aligned}
	f_i^{\rm upper}\!\!=mlp({\bf F}_i)\ \Longrightarrow \
	f_i^{\rm upper}\!\!=mlp({\bf F}_i)(\boldsymbol{W}^{\rm{H_2}})^{\top},
	\end{aligned}
	\end{equation}
where {\small$\boldsymbol{W}^{\rm{H_2}}$} denotes the formula of Architecture 2, which has been introduced in Eqn.~(\ref{3dw}).

\begin{framed}
\vspace{-2pt}
\textbf{Architecture 3} used in \cite{liu2018point2sequence} is added to PointNet++ and RSCNN in order to improve their adversarial robustness and neighborhood consistency.\vspace{-2pt}
\end{framed}
\vspace{-4pt}

Table \ref{tab:comparison} (c) shows details about how to obtain DNNs with Architecture 3.

\begin{framed}
\vspace{-2pt}
\textbf{Architecture 4} designed by \cite{jiang2018pointsift} is added to PointNet++, Point2Sequence, and RSCNN in order to improve their rotation robustness.\vspace{-2pt}
\end{framed}
\vspace{-4pt}

Table \ref{tab:comparison} (d) shows details about how to obtain DNNs with Architecture 4.

\section{Experiments}\label{sec:exp}

To demonstrate the broad applicability of our method, we applied our method to diagnose seven widely used DNNs, including PointNet \cite{qi2017pointnet}, PointNet++ \cite{qi2017pointnet++}, PointConv \cite{wu2019pointconv}, DGCNN \cite{wang2018dynamic}, PointSIFT \cite{jiang2018pointsift}, Point2Sequence \cite{liu2018point2sequence}, and RSCNN \cite{liu2019relation}. These DNNs were trained using three benchmark datasets, including the ModelNet40 dataset \cite{wu20153d}, the ShapeNet\footnote{The ShapeNet dataset for classification is converted from the ShapeNet part segmentation dataset.} dataset \cite{chang2015shapenet}, the 3D MNIST \cite{3dMNIST} dataset.

\begin{table}[t]
	\newcommand{\tabincell}[2]{\begin{tabular}{@{}#1@{}}#2\end{tabular}}
	\begin{center}
		\resizebox{\linewidth}{!}{\begin{tabular}{l|c|c|c|c|c}
				\toprule
				
				\multicolumn{1}{c|}{\multirow{2}{*}{Models}} &\!\!\!\! Information \!\!\!\!&\!\!\!\! Information \!\!\!\!&\!\!\!\! Rotation \!\!\!\!&\!\!\!\! Adversarial \!\!\!\!&\!\!\!\! Neighborhood\!\!\!\!\\
				
				&\!\!\!\! discarding \!\!\!\!&\!\!\!\! concentration \!\!\!\!&\!\!\!\! non-robustness\textcolor{red}{\footnotemark[1]} \!\!\!\!&\!\!\!\! robustness \!\!\!\!&\!\!\!\! inconsistency\textcolor{red}{\footnotemark[1]} \!\!\!\!\\
				
				\midrule
				\!\!\!\! PointNet  \!\!\!\!&\!\!\!\!-8370.42 \!\!\!\!&\!\!\!\!1.089 \!\!\!\!&\!\!\!\!8.000  \!\!\!\!&\!\!\!\! 1.994 \!\!\!\!&\!\!\! 3.127 \!\!\!\!\!\\ 
				
				\!\!\!\! PointNet++  \!\!\!\!&\!\!\!\! -8166.16 \!\!\!\!&\!\!\!\! 1.625 \!\!\!\!&\!\!\!\! 7.093  \!\!\!\!&\!\!\!\! 2.504 \!\!\!\!&\!\!\!\! 3.409\!\!\!\!\\
				
				\!\!\!\! PointConv \!\!\!\!&\!\!\!\! -8766.00 \!\!\!\!&\!\!\!\! 0.865 \!\!\!\!&\!\!\!\! 4.875  \!\!\!\!&\!\!\!\! 2.878 \!\!\!\!&\!\!\!\! 3.781 \!\!\!\!\!\!\\ 
				
				\!\!\!\! DGCNN  \!\!\!\!&\!\!\!\! -9187.51 \!\!\!\!&\!\!\!\! 1.336 \!\!\!\!&\!\!\!\! 5.985  \!\!\!\!&\!\!\!\! 2.421 \!\!\!\!&\!\!\!\! 1.449\!\!\!\!\\ 
				
				\!\!\!\! PointSIFT  \!\!\!\!&\!\!\!\! -8415.92 \!\!\!\!&\!\!\!\! 0.079 \!\!\!\!&\!\!\!\! 3.931  \!\!\!\!&\!\!\!\! 2.839 \!\!\!\!&\!\!\!\! 2.423\!\!\!\!\\ 
				
				\!\!\!\! Point2Sequence  \!\!\!\!&\!\!\!\! -8328.34 \!\!\!\!&\!\!\!\! 1.321 \!\!\!\!&\!\!\!\! 9.506  \!\!\!\!&\!\!\!\! 2.526 \!\!\!\!&\!\!\!\! 3.184\!\!\!\!\\
				
				\!\!\!\! RSCNN \!\!\!\!&\!\!\!\! -8009.52 \!\!\!\!&\!\!\!\! 2.220 \!\!\!\!&\!\!\!\! 3.645  \!\!\!\!&\!\!\!\! 2.314 \!\!\!\!&\!\!\!\! 3.585\!\!\!\!\\
				\bottomrule
		\end{tabular}}
	\end{center}
	\caption{Quantification of the representation capacity of different DNNs on the ModelNet40 dataset.}
	\label{tab:fivemetircs}
\end{table}

\begin{table*}[tbp]
	\begin{center}
		\resizebox{\linewidth}{!}{
			\begin{tabular}{cl|c|c|c|c|c|c|c|c|c}
				\toprule[1pt]
				& &\multicolumn{3}{c|}{ModelNet40 dataset}& \multicolumn{3}{c|}{ShapeNet dataset}& \multicolumn{3}{c}{3D MNIST dataset}\\
				\cline{3-5}\cline{6-8}\cline{9-11}
				&& w/ & w/o & {\small$\Delta$} & w/ & w/o & {\small$\Delta$}  & w/ & w/o & {\small$\Delta$} \\
				\hline
				Hypothesis 1 &Keep/remove Architecture 1 from PointConv for adversarial robustness
				
				&2.878&2.629&{\bf0.249} &2.407 &2.271 &{\bf0.136} &2.737 &2.530 &{\bf0.207}\\
				
				\hline
				Hypothesis 2&Keep/remove Architecture 2 from PointConv for rotation non-robustness
				&4.875	&5.066	&{\bf0.191}	&4.470	&5.368	&{\bf0.898}	&6.650	&8.019	&{\bf1.369} \\
				\hline
				
				\multirow{4}{*}{Hypothesis 3}&Keep/remove Architecture 3 from Point2Sequence (4 \emph{vs.} 3 scales) for adversarial robustness
				
				& \multirow{2}{*}{2.526} & 2.521 & {\bf0.005} & \multirow{2}{*}{2.520} & 2.514 & {\bf0.006} & \multirow{2}{*}{2.468}& 2.479 & -0.011 \\
				
				\cline{2-2} \cline{4-5} \cline{7-8}\cline{10-11}
				
				&\multirow{1}{*}{Keep/remove Architecture 3 from Point2Sequence (4 \emph{vs.} 2 scales) for adversarial robustness}
				
				& & 2.513& {\bf0.013} &   & 2.488 & {\bf0.032} &   & 2.460 & {\bf0.008}  \\
				
				\cline{2-11}
				
				&Keep/remove Architecture 3 from Point2Sequence (4 \emph{vs.} 3 scales) for neighborhood inconsistency
				& \multirow{2}{*}{3.184}&3.332 &{\bf0.148} & \multirow{2}{*}{2.815}&2.992 &{\bf0.177} & \multirow{2}{*}{3.097} &3.342 & {\bf0.245} \\
				
				\cline{2-2} \cline{4-5} \cline{7-8} \cline{10-11}
				
				&Keep/remove Architecture 3 from Point2Sequence (4 \emph{vs.} 2 scales) for neighborhood inconsistency
				&  &3.148  & -0.036 & &2.947  & {\bf0.132} & &3.431  & {\bf0.334} \\

				\hline
				Hypothesis 4&Keep/remove Architecture 4 from PointSIFT for rotation non-robustness
				&3.931	&7.274	&{\bf3.343}	&3.678	&6.223	&{\bf2.545}	&6.308	&5.619	&-0.689\\
				
				\bottomrule[1pt]
		\end{tabular}	}
	\end{center}
	\caption{Verifying hypotheses of utilities of specific intermediate-layer network architectures. The column {\small$\Delta$} denotes the increase of the utility of the network with the specific architecture \emph{w.r.t.} the network without the specific architecture\textcolor{red}{\protect\footnotemark[4]}. In particular, for adversarial robustness, {\small$\Delta$} was calculated as the adversarial robustness of the network w/ the specific architecture minus the adversarial robustness of the network w/o the specific architecture. For rotation non-robustness and neighborhood inconsistency, {\small$\Delta$} was calculated as the rotation non-robustness/neighborhood inconsistency of the network w/o the specific architecture minus the rotation non-robustness/neighborhood inconsistency of the network w/ the specific architecture. \emph{{\small$\Delta>0$} indicates that the corresponding hypothesis has been verified.} Experimental results show that the proposed four hypotheses were verified.}
	\label{tab:hypotheses_verification}
\end{table*}

\begin{table*}[t]
	\begin{center}
		\resizebox{\linewidth}{!}{\begin{threeparttable}
				\begin{tabular}{l|c|c|c|c|c|c|c|c|c}
					\toprule[1pt]
					&\multicolumn{3}{c|}{ModelNet40 dataset}& \multicolumn{3}{c|}{ShapeNet dataset}& \multicolumn{3}{c}{3D MNIST dataset}\\
					\cline{2-4}\cline{5-7}\cline{8-10}
					& added & ori.  & {\small$\Delta$} & added & ori.  & {\small$\Delta$}  & added & ori.  & {\small$\Delta$} \\
					\hline
					
					Add Architecture 1 to PointNet++ for adversarial robustness
					
					&2.519&2.504&{\bf0.015}&2.496&2.437&{\bf0.059}&2.427&2.352& {\bf0.075}\\
					
					Add Architecture 1 to Point2Sequnece for adversarial robustness
					
					&2.544&2.526&{\bf0.018}&2.500&2.520&-0.020&2.475&2.468&{\bf0.007}\\
					
					Add Architecture 1 to RSCNN for adversarial robustness
					
					&2.342&2.314&{\bf0.028}&2.337&2.227&{\bf0.110}&2.283&2.279&{\bf0.004}\\
					
					\hline
					Add Architecture 2 to PointNet++ for rotation non-robustness
					
					&3.845	&7.093	&{\bf3.248}	&2.921	&5.929	&{\bf3.008}	&3.143	&6.531	&{\bf3.388}\\

					Add Architecture 2 to Point2Sequence for rotation non-robustness
					
					&4.963	&9.506	&{\bf4.543}	&4.017	&7.451	&{\bf3.434}	&4.354	&6.890	&{\bf2.536}\\
					
					Add Architecture 2 to RSCNN for rotation non-robustness
					&3.993	&3.645	&-0.348	&4.685	&3.460	&-1.225	&3.391	&3.456	&{\bf0.065}\\
					
					\hline
					
					Add Architecture 3 to PointNet++ for adversarial robustness
					
					&3.010&2.504& {\bf0.506} &2.987&2.437&{\bf0.550}&2.604&2.352&{\bf0.252}\\
					
					Add Architecture 3 to RSCNN for adversarial robustness
					
					&2.350&2.314& {\bf0.036} &2.342&2.279&{\bf0.063}&2.332&2.227&{\bf0.105}\\

					Add Architecture 3 to PointNet++ for neighborhood inconsistency
					&3.010	&3.409	&{\bf0.399}	&3.288	&3.352	&{\bf0.064}	&3.480	&3.541	&{\bf0.062}\\
					
					Add Architecture 3 to RSCNN for neighborhood inconsistency
					&3.497	&3.585	&{\bf0.088}	&3.167	&3.478	&{\bf0.311}	&3.397	&3.928	&{\bf0.531}\\
					
					\hline
					
					Add Architecture 4 to PointNet++ for rotation non-robustness
					&6.191	&7.093	&{\bf0.902}	&4.898	&5.929	&{\bf1.031}	&3.513	&6.531	&{\bf3.018}\\
					
					Add Architecture 4 to Point2Sequence for rotation non-robustness
					&6.005	&9.506	&{\bf3.501}	&8.385	&7.451	&-0.934	&9.494	&6.933	&-2.561\\
					
					Add Architecture 4 to RSCNN for rotation non-robustness
					& 2.424 &3.645	&{\bf1.221}	&2.697	&3.460	&{\bf0.763}	&3.555	&3.456	&-0.099 \\
					\bottomrule[1pt]
				\end{tabular}
		\end{threeparttable}}
	\end{center}
	\caption{Improving utilities of existing DNNs by adding modules of specific intermediate-layer architectures. The column ``added'' denotes the utility of the network which the specific architecture was added to\textcolor{red}{\protect\footnotemark[4]}. The column ``ori.'' denots the utility of the original network. The column {\small$\Delta$} denotes the improvement of the utility of the network which the specific architecture was added to \emph{w.r.t.} the original network. In particular, for adversarial robustness, {\small$\Delta$} was calculated as the value of the ``added'' column minus the value of the ``ori.'' column. For rotation non-robustness and neighborhood inconsistency, {\small$\Delta$} was calculated as the value of the ``ori.'' column minus the value of the ``added'' column. \emph{{\small$\Delta>0$} indicates that the specific architecture improves the utility of the DNN.} Experimental results show that the verified hypotheses could be used to revise existing DNNs to improve their utilities.}
	\label{tab:utility}
\end{table*}

\textbf{Implementation details:} When computing the information discarding, we bounded each dimension of {\small$\boldsymbol{\sigma} = [\sigma_1,\sigma_2,\dots]^{\top}$} within 0.08 for fair comparison between different DNNs. It was because when processing a point cloud, some widely used operations (\emph{e.g.} the {\small$g(\cdot)$} operation in Eqn. (\ref{ms})) would randomly and completely discard the information of some points. This resulted in that the learned {\small$\sigma$} corresponding to these points could be infinite in theory. For the computation of rotation robustness, during the training and testing phases, each point cloud was rotated by random angles. For the computation of neighborhood inconsistency, we used {\small$k$}-NN search to select 16 neighbors for each point.

To analyze the information concentration of DNNs, we generated a new dataset that contained both the foreground objects and the background, since most widely used benchmark datasets for point cloud classification only contain foreground objects. Specifically, for each sample (\emph{i.e.} the foreground object) in the ModelNet40, we generated the background as follows. First, we randomly sampled a set of 500 points from point clouds, which had different labels from the foreground object. Second, we resized this set of points to the density of the foreground object. Finally, we randomly located it around the foreground object. \emph{The dataset will be released when this paper is accepted.}

The entropy-based method \cite{ma2019quantifying} quantified the layerwise information discarding. This method assumed the feature space of the concept of a specific object satisfied {\small$\left\|f'-f\right\|^2 < \epsilon$}, where {\small$f=h(X)$}, {\small$f'=h(X')$}, {\small$X'=X+\boldsymbol{\delta}$}. {\small$\boldsymbol{\delta}$} denotes a random noise. For point cloud processing, each dimension of the intermediate-layer feature is computed using the context of a specific point {\small$x_i$}. However, adding noise to a point cloud will change the context of each point. In order to extend the entropy-based method to point cloud processing, we selected the same set of points as the contexts \emph{w.r.t.} {\small$x_i$} and {\small$x'_i$}, so as to generate a convincing evaluation\footnote{Detailed discussions are presented in supplementary materials.}.

\textbf{Comparisons of the representation capacity of DNNs:} As shown in Table~\ref{tab:fivemetircs}, we measured the proposed metrics for the fully connected layer close to the network output\footnote{Theoretically, features of any layer can be used to analyze the representation capacity of DNNs.}, which had 512 hidden units. We measured adversarial robustness by performing adversarial attacks over all incorrect classes. We found that PointNet++ and RSCNN had relatively higher values of information discarding. PointNet++ and RSCNN concentrated more on the foreground object. PointConv, DGCNN, PointSIFT, and RSCNN performed well in rotation robustness. PointConv and PointSIFT exhibited higher adversarial robustness. DGCNN and PointSIFT exhibited lower neighborhood inconsistency.

\textbf{Verifying hypotheses of utilities of specific network intermediate-layer architectures:} As shown in Table~\ref{tab:hypotheses_verification}, the proposed four hypotheses had been verified. Architecture 1 improved the utility of adversarial robustness. One possible reason was that Architecture 1 considered distances between each point and its neighbors during the computing of densities, which increased the difficulty of adversarial attacks. Architecture 3 also improved the utility of adversarial robustness. We found that the utility of adversarial robustness increased with the scale number of features. The reason may be that concatenating features with different scales enhanced the representation capacity, so that it was more challenging to conduct adversarial attacks. Architecture 2 and Architecture 4 improved the utility of rotation robustness. The reason may be that both Architecture 2 and Architecture 4 extracted contextual information from coordinates of each point's neighbors using non-linear transformations. Such contextual information improved rotation robustness. Besides, networks with Architecture 3 usually had lower neighborhood inconsistency than those without Architecture 3. DNNs that extracted features from contexts of more scales usually exhibited lower neighborhood inconsistency. One possible reason was that extracting multi-scale features enhanced connections between neighboring points.

\textbf{Improving utilities of existing DNNs:} In this experiment, we aimed to prove that utilities of existing DNNs could be improved by using the verified hypotheses to guide the architectural revision. To this end, we conducted comparative studies as designed in Table \ref{tab:comparison}. As shown in Table \ref{tab:utility}, adding specific intermediate-layer architectures to existing DNNs improved their utilities. Specifically, adding modules of Architecture 1 improved the utility of adversarial robustness of PointNet++, Point2Sequence, and RSCNN. Adding modules of Architecture 2 improved the utility of rotation robustness of PointNet++, Point2Sequence, and RSCNN. Adding modules of Architecture 3 improved utilities of adversarial robustness and neighborhood consistency of both PointNet++ and RSCNN. Adding modules of Architecture 4 improved the utility of rotation robustness of PointNet++, Point2Sequence, and RSCNN.

\textbf{Relationship between utilities and accuracy:} Note that this study focused on the verification of utilities of specific network architectures and the architectural revision of existing DNNs to improve their utilities, instead of the classification accuracy. Nevertheless, we provided the classification accuracy of different versions of DNNs in supplementary materials. \emph{Experimental results show that adding a specific architecture to existing DNNs has effects on accuracy. We have detailedly discussed the relationship between utilities and accuracy in supplementary materials.}

\section{Conclusion}

In this paper, we have verified a few hypotheses of the utility of four specific intermediate-layer network architectures for 3D point cloud processing. Comparative studies are conducted to prove the utility of the specific architectures, including rotation robustness, adversarial robustness, and neighborhood inconsistency. In preliminary experiments, we have verified that Architecture 2 and Architecture 4 mainly improve the rotation robustness; Architecture 1 and Architecture 3 have positive effects on adversarial robustness; Architecture 3 usually alleviates the neighborhood inconsistency. These verified hypotheses have further been used to revise existing DNNs to improve their utilities.

Considering that unnecessarily complex intermediate-layer architectures will bring in uncertainty of experiments, we only verify utilities of simple network architectures \emph{w.r.t.} the object classification. More generic hypotheses about utilities of other tasks (\emph{e.g.} segmentation and reconstruction) need to be verified in the future.

~\\
\noindent
\textbf{Acknowledgments} This work is partially supported by the National Nature Science Foundation of China (No. 61906120, U19B2043, 61976160).

{\small
\bibliographystyle{ieee_fullname}
\bibliography{egbib}
}

\onecolumn
\appendix

\section{Overview}
This Appendix provides more details about comparative studies in the main paper and includes more implementation details about experiments. In Section~\ref{appendix:max}, we introduce a special element-wise max operator widely used in point cloud processing. In Section~\ref{appendix:dnns}, we briefly introduce DNNs used in comparative studies. In Section~\ref{appendix:versions}, we show details about different versions of DNNs for comparison. In Section~\ref{appendix:fix}, we show implementation details about extending the entropy-based method \cite{ma2019quantifying} to point cloud processing. In Section~\ref{appendix:accuracy}, we compare the accuracy of different versions of DNNs. In Section~\ref{appendix:relatedwork}, we supplement related work about learning interpretable representations.

\section{Details about the function $\boldsymbol{g}(\cdot)$}\label{appendix:max}

\begin{figure*}[h]
	\centering
	\includegraphics[width=0.4\linewidth]{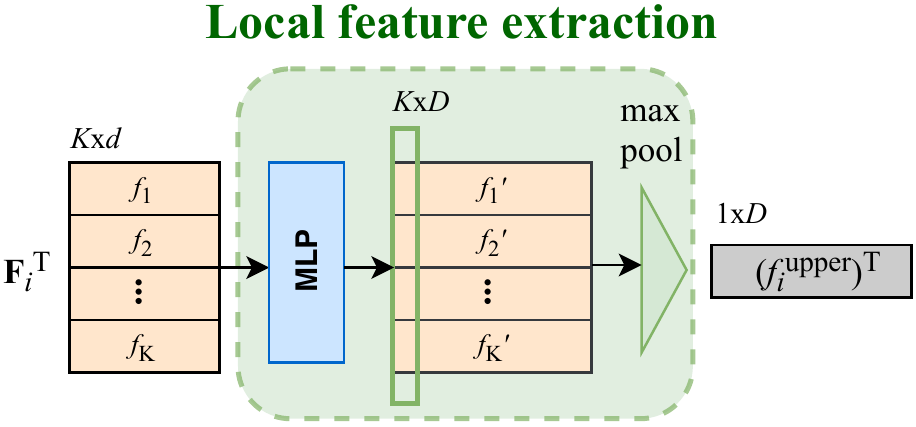}
	\caption{\textbf{Illustration of the special max pooling operator {\small$\boldsymbol{g}(\cdot)$}.} {\small${\bf F}_i$} denotes features correspond to points in neighborhood {\small$\mathbf{N}(i)$} \emph{w.r.t.} point {\small$x_i$}. Each row of {\small$\mathbf{F}_i^{\top}$} in the figure represents the feature of a specific point in {\small$\mathbf{N}(i)$}.  }
	\label{fig:max}
\end{figure*}

In point cloud processing, a special element-wise max operator, {\small${g}(\cdot)$}, is widely used for aggregating a set of neighboring points' features into a local feature. As shown in Figure~\ref{fig:max}, given a set of {\small$K$} nearest neighboring points of {\small$x_i$}, {\small$\mathbf{N}(i)$}, let {\small$\mathbf{F}_i\in \mathbb{R}^{d\times K}$} denote intermediate-layer features that correspond to the set of neighboring points in {\small$\mathbf{N}(i)$} \emph{w.r.t.} the point {\small$x_i$}. Each specific column of {\small$\mathbf{F}_i$} represents the feature of a specific point in {\small$\mathbf{N}(i)$}. The feature in the upper layer, \emph{i.e.} {\small$f_i^{\rm upper}$}, which is the local feature of {\small$\mathbf{N}(i)$}, can be formulated as follows.
\begin{small}
	\begin{equation}\label{flocal}
	f_i^{\rm upper}=g({\bf F}_i)=\mathop{\mathbf{MAX}}_{i=1,\dots,K}(MLP({\bf F}_i)),
	\end{equation}
\end{small}
where {\small$MLP$} is an MLP with a few layers; {\small$MLP({\bf F}_i)\in\mathbb{R}^{D\times K}$}; {\small$\textbf{MAX}$} is an element-wise max operator as follows. Let {\small${\bf F}'_i =MLP({\bf F}_i)$}.
\begin{small}
	\begin{equation}\label{maxpool}
	\mathop{\mathbf{MAX}}_{i=1,\dots,K}({\bf F}'_i)=\mathop{\mathbf{MAX}}_{i=1,\dots,K}
	\begin{bmatrix}
	f'_{11}      & \cdots & f'_{1K}      \\
	\vdots & \ddots & \vdots \\
	f'_{D1}      & \cdots & f'_{DK}
	\end{bmatrix}\xlongequal{\textbf{define}}
	<\max\limits_{k=1,\dots,K}f'_{1k},\dots,\max\limits_{k=1,\dots,K}f'_{Dk}>^{\top}
	\end{equation}
\end{small}

\section{Summaries of relevant technologies in existing DNNs}\label{appendix:dnns}
For the convenience of readers to quickly understand relevant technologies in existing DNNs, we summarize relevant technologies of PointNet++ \cite{qi2017pointnet++}, PointConv \cite{wu2019pointconv}, Point2Sequence \cite{liu2018point2sequence}, PointSIFT \cite{jiang2018pointsift}, and RSCNN \cite{liu2019relation} in this section.

\subsection{PointNet++}\label{appendix:intro-plus}
PointNet++ \cite{qi2017pointnet++} is a hierarchical structure composed of a number of \emph{set abstraction} modules (SA module). For each SA module, a set of points is processed and abstracted to produce a new set with fewer elements. An SA module includes four parts: the \emph{Sampling layer}, the \emph{Grouping layer}, the \emph{MLP}, and the \emph{Maxpooling layer}. Given a set of {\small$N$} input points, the \emph{Sampling layer} uses the farthest point sampling algorithm to select a subset of points from the input points, which defines the centroids of local regions, {\small$\{x_i\},i=1,\dots,N'$}. Then, for each selected point, the \emph{Grouping layer} constructs a local region by using ball query search to find {\small$K$} neighboring points within a radius {\small$r$}. For each local region {\small$\mathbf{N}(i)$} centered at {\small$x_i$}, {\small${\bf F}_i\in\mathbb{R}^{d\times K}$} denotes the intermediate-layer features that correspond to points in {\small$\mathbf{N}(i)$}. The \emph{MLP} transforms {\small${\bf F}_i$} into higher dimensional features {\small${\bf F}'_i\in\mathbb{R}^{D\times K}$}, where {\small$D>d$}. Finally, the \emph{Maxpooling layer} encodes {\small${\bf F}'_i$} into a local feature {\small$f_i^{\rm upper}$}, which will be fed to the upper SA module. Please see Section~\ref{appendix:max} for details about the \emph{Maxpooling layer}.

In this study, the baseline network of PointNet++ is composed of three SA modules and a few fully connected layers. Please see Table~\ref{tab:plus1} (left column) for details about the network architecture.

\subsection{PointConv}\label{appendix:intro-conv}

PointConv \cite{wu2019pointconv} has a similar architecture with PointNet++, \emph{i.e.} hierarchically using a few blocks to extract contextual information. In this study, the baseline network of PointConv is composed of five blocks. Each block is constructed as [\emph{Sample layer}$\to$\emph{Group layer}$\to$\emph{MLP}$\to$Architecture 1$\to$Architecture 2$\to$\emph{Conv layer}].

The \emph{Sampling layer} uses the farthest point sampling algorithm to select a subset of points from the input points, which defines the centroids of local regions. Then, for each selected point, the \emph{Grouping layer} constructs a local region by using {\small$k$}-NN search to find {\small$K$} neighboring points. For each local region, the \emph{MLP} transforms features of points in the local region into higher dimensional features. Different from PointNet++, PointConv uses the information of density (\emph{i.e.} Architecture 1) and local 3D coordinates (\emph{i.e.} Architecture 2) to reweight the features learned by the \emph{MLP}. Finally, a {\small$1\times 1$} convolution is used to compute the output feature of each local region. Please see Table~\ref{tab:conv1} (left column) for details about the network architecture.

\subsection{Point2Sequence}\label{appendix:intro-p2s}
Point2Sequence \cite{liu2018point2sequence} is composed of five parts: (a) multi-scale area establishment, (b) area feature extraction, (c) encoder-decoder feature aggregation, (d) local region feature aggregation, and (e) shape classification, where parts (a) and (b) make up Architecture 3 in our study.

Specifically, given a point cloud {\small $\mathbb{X}=\{x_i\},i=1,2,...,N$}, Point2Sequence first uses the farthest point sampling algorithm to select $N'$ points from the input point cloud, {\small$\mathbb{X}'=\{x'_j\},j=1,2,...,N'$}, to define the centroids of local regions {\small$\{\mathbf{N}(j)\},j=1,2,...N'$}. For each local region {\small$\mathbf{N}(j)$}, {\small$T$} different scale areas {\small$\{\mathbf{A}(j)^1,..., \mathbf{A}(j)^t,...,\mathbf{A}(j)^T\}$ } are established by using {\small$k$}-NN search to select {\small$T$} nearest points of {\small$x'_j$}, {\small$[K_1,...,K_t,...,K_T]$}. In this way, multi-scale areas are established. Then, Point2Sequence extracts a feature {\small$f_{j,{\rm scale}=K_t}^{\rm upper}\in\mathbb{R}^d$} for each scale area {\small$\mathbf{A}(j)^t$} by the \emph{MLP} and the \emph{Maxpooling layer} introduced in Section~\ref{appendix:intro-plus}. In this way, for each local region {\small$\mathbf{N}(j)$}, a feature sequence {\small$f_j^{\rm upper}=\{f_{j,{\rm scale}=K_1}^{\rm upper},...,f_{j,{\rm scale}=K_t}^{\rm upper},...,f_{j,{\rm scale}=K_T}^{\rm upper}\}$} is obtained. Then, {\small$f_j^{\rm upper}$} is aggregated into a {\small$d$}-dimensional feature {\small$\mathbf{r}_j$} by the encoder-decoder feature aggregation part. The sequence encoder-decoder structure used here is an LSTM network, where an attention mechanism is proposed to highlight the importance of different area scales (please see \cite{liu2018point2sequence} for details). Then, a 1024-dimensional global feature is aggregated from the features {\small$\mathbf{r}_j$} of all local regions by the local region feature aggregation part. Finally, the global feature is used for shape classification. Please see Table~\ref{tab:p2s_ori} for details about the network architecture.

\subsection{PointSIFT}\label{appendix:intro-oe}
PointSIFT \cite{jiang2018pointsift} adopts the similar hierarchical structure as PointNet++, which is composed of a number of SA modules. The difference is that PointSIFT uses a special orientation encoding unit, \emph{i.e.}, Architecture 4, to learn an orientation-aware feature for each point.

Architecture 4 is a point-wise local feature descriptor that encodes information of eight orientations. Unlike the unordered operator, \emph{e.g.} \emph{max pooling}, which discards all inputs except for the maximum, Architecture 4 is an ordered operator, which could be more informative.

Architecture 4 first selects 8-nearest points of {\small$x_i$} from eight octants partitioned by the ordering of three coordinates. Since distant points provide little information for the description of local patterns, when no point exists within searching radius {\small$r$} in some octant, {\small$x_i$} will be duplicated as the nearest neighbor of itself. Then, Architecture 4 processes features of 8-nearest neighboring points, {\small${\bf F}_i^{\rm oe} \in \mathbb{R}^{d\times 2 \times 2 \times 2 }$}, which reside in a {\small$2\times 2 \times 2$} cube for local pattern description centering at {\small$x_i$}, the three dimensions {\small$2 \times 2 \times 2$} correspond to three axes. An orientation-encoding convolution, \emph{i.e.} {\small$Conv^{\rm oe}$}, which is a three-stage operator, is used to convolve the {\small$2\times 2 \times 2$} cube along {\small$\rm{x}$}, {\small$\rm{y}$}, and {\small$\rm{z}$} axis. The three-stage convolution {\small$Conv^{\rm oe}$} is formulated as:
\begin{small}
	\begin{equation}
	\begin{aligned}
	f_i^{\rm{x}-axis} &= ReLU(Conv(W_{\rm{x}},{\bf F}_i^{\rm oe}))) \in \mathbb{R}^{d\times 2\times 2\times 1}\\
	f_i^{\rm{(x,y)}-axis}&= ReLU(Conv(W_{\rm{y}},f_i^{\rm{x}-axis})) \in \mathbb{R}^{d\times 2\times 1\times 1}\\
	f_i^{\rm oe}=f_i^{\rm{(x,y,z)}-axis}&= ReLU(Conv(W_{\rm{z}},f_i^{\rm{(x,y)}-axis}))\in \mathbb{R}^{d\times 1\times 1\times 1}
	\end{aligned}
	\end{equation}
\end{small}
where {\small$W_{\rm{x}}\in \mathbb{R}^{d\times 1\times 1\times 2}$}, {\small$W_{\rm{y}}\in \mathbb{R}^{d\times 1\times 2\times 1}$}, and {\small$W_{\rm{z}}\in \mathbb{R}^{d\times 2\times 1\times 1}$} are weights of the convolution operator.

In this way, Architecture 4 learns the orientation-aware feature {\small$f_i^{\rm oe}$} for each point {\small$x_i$}. Such orientation-aware features will be fed to SA modules (introduced in Section~\ref{appendix:intro-plus}) to extract contextual information. Please see Table~\ref{tab:siftw/ooe} (left column) for details about the network architecture.

\begin{table*}[tbp]
	\centering
	\resizebox{0.8\linewidth}{!}{\begin{tabular}{c|c|c|c}
			\hline
			Pointnet++             & Pointnet++ with Architecture 1        & Pointnet++ with Architecture 2        & Pointnet++ with Architecture 4                  \\ \hline
			Sample (512)           & Sample (512)                          & Sample (512)  & Sample (512)            \\
			Group (0.2,32)         & Group (0.2,32)                        & Group (0.2,32)                        & Group (0.2,32)                                  \\
			MLP [64,64,128{]}    & MLP [64,64,128]                  & MLP [64,64,128]                & MLP [64,64,128]                           \\
			Maxpooling             & \textbf{Architecture 1} & \textbf{ Architecture 2} & Maxpooling                                      \\
			Sample (128)           & Maxpooling                            & Maxpooling                            & Sample (128)                                    \\
			Group (0.4,64)         & Sample (128)                          & Sample (128)                          & Group (0.4,64)          \\
			MLP [128,128,256]  & Group (0.4,64)                        & Group (0.4,64)                        & MLP [128,128,256] \\
			Maxpooling             & MLP [128,128,256]                & MLP [128,128,256]                 & Maxpooling                                      \\
			Sample (1)             & \textbf{Architecture 1} & \textbf{ Architecture 2} & \textbf{Architecture 4 [256]} \\
			Group (all)            & Maxpooling     &  Maxpooling     & Sample (1)                                      \\
			MLP [256,512,1024]& Sample (1)                            & Sample (1)                            & Group (all)           \\
			Maxpooling             & Group (all)                           & Group (all)                           & MLP [256,512,1024]  \\
			FC [512,256,40]    & MLP [256,512,1024]               & MLP [256,512,1024]                & Maxpooling                                      \\
			Softmax                & \textbf{Architecture 1} & \textbf{ Architecture 2} & FC [512,256,40]                             \\
			& Maxpooling                            & Maxpooling                            & Softmax                                         \\
			& FC [512,256,40]                   & FC [512,256,40]                  &                                                 \\
			& Softmax                               & Softmax                               &                                                 \\ \hline
	\end{tabular}}
	\caption{Different versions of PointNet++, including the original one, the one with Architecture 1, the one with Architecture 2, and the one with Architecture 4. Sample ({\small$N$}) indicates the \emph{Sample layer}, which selects a subset of {\small$N$} points from the input point cloud. Group ({\small$r,K$}) indicates the \emph{Group layer}, which uses the ball query search to find {\small$K$} neighboring points around each sampled point within a radius {\small$r$}. Group (all) means constructing a region with all the input points. MLP {\small$[u_1,\dots,u_l]$} indicates the MLP with {\small$l$} layers, where {\small$u_i$} is the number of hidden units of the {\small$i$}-th layer. Architecture 4 [d] indicates Architecture 4, which outputs {\small$d$}-dimensional features.}
	\label{tab:plus1}
\end{table*}

\subsection{RSCNN}\label{appendix:intro-rscnn}
RSCNN \cite{liu2019relation} adopts the similar hierarchical structure as PointNet++, which is composed of a number of SA modules. The difference is that RSCNN uses a special \emph{relation-shape convolution} (RS-Conv) to learn from the relation, \emph{i.e.} \emph{the geometric topology constraint among points. Specifically, the convolutional weight for local point set is forced to learn a high-level relation expression from predefined geometric priors, between a sampled point from this point set and the others.}

The goal of the RS-Conv operation is to learn an inductive representation of the neighborhood of each point. Given a point {\small$x_i$}, let {\small$\mathcal{N}(x_i)$} be the neighborhood centered at {\small$x_i$}. Each point {\small$x_j\in\mathcal{N}(x_i)$} is the surrounding point of {\small$x_i$}. The RS-Conv operation consists of two steps: (1) learning from relation, and (2) channel-raising mapping. The first step can be formulated as follows.
\begin{small}
	\begin{equation}
	\mathbf{f}_{P_{\rm sub}} = \sigma(\mathcal{A}(\{MLP(\mathbf{h}_{ij})\cdot\mathbf{f}_{x_j}, \forall x_j \}) ) ,\quad d_{ij}<r\ \ \forall x_j\in\mathcal{N}(x_i),
	\end{equation}
\end{small}
where $d_{ij}$ is the Euclidean distance between $x_i$ and $x_j$ , and $r$ is the sphere radius.

The step of \emph{learning from relation} can be summarized as follows. First, transform features of all the points in {\small$\mathcal{N}(x_i)$} with function {\small$MLP(\mathbf{h}_{ij})\cdot\mathbf{f}_{x_j}$}, where {\small$MLP(\mathbf{h}_{ij})$} uses a shared MLP to abstract high-level relation expression between points {\small$x_i$} and {\small$x_j$} and {\small$\mathbf{h}_{ij}$} is defined as a compact vector with 10 channels, \emph{i.e.} (3D Euclidean distance, {\small$x_i-x_j$}, {\small$x_i$}, {\small$x_j$}). Then, aggregate the transformed features with function $\mathcal{A}$ followed by a nonlinear activator {\small$\sigma$}. {\small$\sigma$} is implemented as the special maxpooling operation introduced in Section~\ref{appendix:max}.

The step of \emph{channel-raising mapping} is to increase the channel number of {\small$\mathbf{f}_{P_{\rm sub}}$}. Specifically, a shared MLP is added on {\small$\mathbf{f}_{P_{\rm sub}}$} to achieve the goal. Please see Table~\ref{tab:rscnn1} (left column) for details about the network architecture.

\section{Global architectures of existing DNNs and their revised versions}\label{appendix:versions}

\begin{table}[tbp]
	\centering
	\resizebox{0.6\linewidth}{!}{\begin{tabular}{c|c|c|c}
			\hline
			PointNet++             & \multicolumn{3}{c}{PointNet++ with Architecture 3}                                                                                              \\ \hline
			Sample (512)           & \multicolumn{3}{c}{Sample (512)}                                                                                                           \\ \cline{2-4}
			Group (0.2,32)         & \textbf{Group (0.1,16)}      & Group (0.2,32)                               &\textbf{Group (0.4,128)}       \\
			MLP {[}64,64,128{]}    &\textbf{MLP {[}32,32,64{]}}  & MLP {[}64,64,128{]}                          & \textbf{MLP {[}64,96,128{]}}   \\
			Maxpooling             & \textbf{Maxpooling}           & Maxpooling                                    &\textbf{ Maxpooling}             \\ \cline{2-4}
			Sample (128)           & \multicolumn{3}{c}{Multi-Scale Feature Aggregation}                                                                                        \\
			Group (0.4,64)         & \multicolumn{3}{c}{{ Sample (128)}}                                                                                    \\ \cline{2-4}
			MLP {[}128,128,256{]}  & \textbf{Group (0.2,32)}      & { Group (0.4,64)}        &\textbf{Group (0.8,128)}       \\
			Maxpooling             & \textbf{MLP {[}64,64,128{]}} & { MLP {[}128,128,256{]}} &\textbf{MLP {[}128,128,256{]}} \\
			Sample (1)             & \textbf{Maxpooling}           & { Maxpooling}             & \textbf{Maxpooling}             \\ \cline{2-4}
			Group (all)            & \multicolumn{3}{c}{{ Multi-Scale Feature Aggregation}}                                                                 \\
			MLP {[}256,512,1024{]} & \multicolumn{3}{c}{{ Sample (1)}}                                                                                      \\
			Maxpooling             & \multicolumn{3}{c}{{ Group (all)}}                                                                                     \\
			FC {[}512,256,40{]}    & \multicolumn{3}{c}{MLP {[}256,512,1024{]}}                                                                                                 \\
			Softmax                & \multicolumn{3}{c}{FC {[}512,256,40{]}}                                                                                                    \\
			& \multicolumn{3}{c}{Softmax}                                                                                                                \\ \hline
	\end{tabular}}
	\caption{The original PointNet++ and the PointNet++ with Architecture 3.}
	\label{tab:plus2}
\end{table}

\begin{table}[htbp]
	\centering
	\resizebox{0.6\textwidth}{!}{
		\begin{tabular}{ccc}
			\hline
			PointConv                             & PointConv without Architecture 1          & PointConv without Architecture 2          \\ \hline
			Sample (1024)                         & Sample (1024)                         & Sample (1024)                         \\
			Group (32)                            & Group (32)                            & Group (32)                            \\
			MLP [32,32]         & MLP [32,32]       & MLP [32,32]                      \\
			\textbf{Architecture 1} &\textbf{Architecture 2} & \textbf{Architecture 1} \\
			\textbf{Architecture 2} & Conv [64]   & Conv [64]          \\
			Conv [64] & Sample (256)                          & Sample (256)                          \\
			Sample (256)                          & Group (32)                            & Group (32)                            \\
			Group (32)           & MLP [64,64]       & MLP [64,64]                     \\
			MLP [64,64]  & \textbf{Architecture 2} & \textbf{Architecture 1} \\
			\textbf{Architecture 1} & Conv [128]  & Conv [128]    \\
			\textbf{Architecture 2} & Sample (64)                           & Sample (64)                           \\
			Conv [128]     & Group (32)                            & Group (32)                            \\
			Sample (64)                           & MLP [128,128] & MLP [128,128]                \\
			Group (32)                            & \textbf{Architecture 2} & \textbf{Architecture 1} \\
			MLP [128,128]   & Conv [256]  & Conv [256]     \\
			\textbf{Architecture 1} & Sample (36)                           & Sample (36)                           \\
			\textbf{Architecture 2} & Group (32)                            & Group (32)                            \\
			Conv [256] & MLP [256,256]   & MLP [256,256]    \\
			Sample (36)                           &\textbf{Architecture 2} & \textbf{Architecture 1} \\
			Group (32)                            & Conv [512]   & Conv [512]          \\
			MLP [256,256] & Sample (1)                            & Sample (1)                            \\
			\textbf{Architecture 1} & Group (all)                           & Group (all)                           \\
			\textbf{Architecture 2} & MLP [512,512]        & MLP [512,512]           \\
			Conv [512]     & \textbf{Architecture 2} & \textbf{Architecture 1} \\
			Sample (1)                            & Conv {[}1024{]}                        & Conv {[}1024{]}                        \\
			Group (all)                           & FC {[}512,128,40{]}                   & FC {[}512,128,40{]}                   \\
			MLP {[}512,512{]}                     & Softmax                               & Softmax                               \\
			\textbf{Architecture 1} &                                       &                                       \\
			\textbf{Architecture 2} &                                       &                                       \\
			Conv {[}1024{]}                        &                                       &                                       \\
			FC {[}512,128,40{]}                   &                                       &                                       \\
			Softmax                               &                                       &                                       \\ \hline
	\end{tabular}}
	\caption{Different versions of PointConv, including the original one, the one without Architecture 1, the one without Architecture 2. Here Group ({\small$K$}) indicates the \emph{Group layer}, which uses the {\small$k$}-NN search to find {\small$K$} neighboring points around each sampled point. }
	\label{tab:conv1}
\end{table}

\subsection{PointNet++}\label{appendix:plus}

In this study, we reconstructed the PointNet++ \cite{qi2017pointnet++} using four specific modules. Table~\ref{tab:plus1} and Table~\ref{tab:plus2} compare the different versions of PointNet++, including the original one, the one with Architecture 1 \cite{wu2019pointconv}, the one with Architecture 2 \cite{wu2019pointconv}, the one with Architecture 4 \cite{jiang2018pointsift}, and the one with Architecture 3 \cite{liu2018point2sequence}.

To obtain the PointNet++ with Architecture 1 (as shown in Table~\ref{tab:plus1}), we added modules of Architecture 1 after all the \emph{MLPs} in PointNet++, \emph{i.e.} the output of the \emph{MLP} was reweighted by the weights learned by Architecture 1. Architecture 1 used in this study was an MLP with two layers, the first layer contained 16 hidden units, and the second layer contained 1 hidden unit. This network was designed to verify the effect of Architecture 1 on the adversarial robustness.

To obtain the PointNet++ with Architecture 2 (as shown in Table~\ref{tab:plus1}), we added modules of Architecture 2 after all the \emph{MLPs} in PointNet++, \emph{i.e.} the output of the \emph{MLP} was reweighted by the weights learned by Architecture 2. Architecture 2 used in this study was an MLP with a single-layer, which contained 32 hidden units. This network was designed to verify the effect of Architecture 2 on the rotation robustness.

To obtain the PointNet++ with Architecture 4 (as shown in Table~\ref{tab:plus1}), we added the module of Architecture 4 before the last \emph{Sample layer} in PointNet++. This network was designed to verify the effect of Architecture 4 on the rotation robustness.

To obtain the PointNet++ with Architecture 3 (as shown in Table~\ref{tab:plus2}), we used the multi-scale version of PointNet++ designed in \cite{qi2017pointnet++}. Compared with the single-scale version of PointNet++ (as shown in Table~\ref{tab:plus1} (left)), the multi-scale version added two blocks after the first \emph{Sample layer}, \emph{i.e.} {\small$[\emph{Group} (16)\to \emph{MLP} [32,32,64]\to \emph{Maxpooling}]$} and {\small$[\emph{Group} (128)\to \emph{MLP} [64,96,128]\to \emph{Maxpooling}]$}, The multi-scale version added another two blocks after the second \emph{Sample layer}, \emph{i.e.} {\small$[\emph{Group} (32)\to \emph{MLP} [64,64,128]\to \emph{Maxpooling}]$} and {\small$[\emph{Group} (128)\to \emph{MLP} [128,128,256]\to \emph{Maxpooling}]$}. In this way, the multi-scale version of PointNet++ extracted two more scale features. This network was used to verify the effect of Architecture 3 on the adversarial robustness and the neighborhood inconsistency.

\subsection{PointConv} \label{appendix:pointconv}
Table~\ref{tab:conv1} compares different versions of PointConv \cite{wu2019pointconv}, including the original one, the one without Architecture 1 \cite{wu2019pointconv}, and the one without Architecture 2 \cite{wu2019pointconv}.

To obtain the PointConv without Architecture 1 (as shown in Table~\ref{tab:conv1} (middle column)), we removed all the five modules of Architecture 1 from the original PointConv architecture. This network was designed to verify the effect of Architecture 1 on the adversarial robustness.

To obtain the PointConv without Architecture 2 (as shown in Table~\ref{tab:conv1} (right column)), we removed all the five modules of Architecture 2 from the original PointConv architecture. This network was designed to verify the effect of Architecture 2 on the rotation robustness.

\subsection{Point2Sequence} \label{appendix:p2s}

The baseline network of Point2Sequence (as shown in Table~\ref{tab:p2s_ori}) extracted features of four different scales, \emph{i.e.}, for each local region centered at point {\small$x_i$}, features were computed using the contextual information of 16, 32, 64, and 128 nearest neighbors of {\small$x_i$}, respectively. To obtain different versions of Point2Sequence for comparison, we removed features of specific scales. We first removed the feature extracted by {\small$[\emph{Group} (16)\to \emph{MLP} [32,64,128]\to\emph{Maxpooling}]$} to obtain the first version of Point2Sequence. We then removed features extracted by {\small$[\emph{Group} (16)\to \emph{MLP} [32,64,128]\to\emph{Maxpooling}]$} and {\small$[\emph{Group} (32)\to \emph{MLP} [64,64,128]\to\emph{Maxpooling}]$} to obtain the second version for comparison. These two versions for comparison were designed to verify the effect of Architecture 3 on the adversarial robustness and the neighborhood inconsistency.

\begin{table}[ht]
	\centering
	\resizebox{0.55\textwidth}{!}{%
		\begin{tabular}{cccc}
			\hline
			\multicolumn{4}{c}{Point2Sequence}                                                                                                                                                                       \\ \hline
			\multicolumn{4}{c}{Sample (384)}                                                                                                                                                                         \\ \hline
			\multicolumn{1}{c|}{{ Group (16)}}          & \multicolumn{1}{c|}{Group (32)}          & \multicolumn{1}{c|}{{Group (64)}}          & Group4 (128)           \\
			\multicolumn{1}{c|}{{ MLP {[}32,64,128{]}}} & \multicolumn{1}{c|}{MLP {[}64,64,128{]}} & \multicolumn{1}{c|}{{ MLP {[}64,64,128{]}}} & MLP4 {[}128,128,128{]} \\
			\multicolumn{1}{c|}{{ Maxpooling}}           & \multicolumn{1}{c|}{Maxpooling}           & \multicolumn{1}{c|}{{ Maxpooling}}           & Maxpooling             \\ \hline
			\multicolumn{4}{c}{Multi-Scale Feature Aggregation}                                                                                                                                                      \\
			\multicolumn{4}{c}{{ LSTM {[}128{]}}}                                                                                                                                                \\
			\multicolumn{4}{c}{{ Sample (1)}}                                                                                                                                                    \\
			\multicolumn{4}{c}{{ Group (all)}}                                                                                                                                                   \\
			\multicolumn{4}{c}{MLP {[}256,512,1024{]}}                                                                                                                                                               \\
			\multicolumn{4}{c}{Maxpooling}                                                                                                                                                                           \\
			\multicolumn{4}{c}{FC {[}512,256,40{]}}                                                                                                                                                                  \\
			\multicolumn{4}{c}{Softmax}                                                                                                                                                                              \\ \hline
	\end{tabular}	}
	\caption{Illustration of the original Point2Sequnence network architecture. Here Group ({\small$K$}) indicates the \emph{Group layer}, which uses the {\small$k$}-NN search to find {\small$K$} neighboring points around each sampled point.}
	\label{tab:p2s_ori}
\end{table}

To obtain the Point2Sequence with Architecture 1 (as shown in Table~\ref{tab:p2sw/density}), we added the module of Architecture 1 after the last \emph{MLP}, \emph{i.e. MLP} [256,512,1024], in Point2Sequence. This network was designed to verify the effect of Architecture 1 on the adversarial robustness.

\begin{table}[htbp]
	\centering
	\resizebox{0.55\textwidth}{!}{%
		\begin{tabular}{cccc}
			\hline
			\multicolumn{4}{c}{Point2Sequence with Architecture 1}                                                                                                                                                                                   \\ \hline
			\multicolumn{4}{c}{Sample (384)}                                                                                                                                                                                           \\ \hline
			\multicolumn{1}{c|}{Group (16)}                        & \multicolumn{1}{c|}{Group (32)}                        & \multicolumn{1}{c|}{{ Group (64)}}          & Group4 (128)                        \\
			\multicolumn{1}{c|}{MLP {[}32,64,128{]}}               & \multicolumn{1}{c|}{MLP {[}64,64,128{]}}               & \multicolumn{1}{c|}{{ MLP {[}64,64,128{]}}} & MLP4 {[}128,128,128{]}              \\
			\multicolumn{1}{c|}{Maxpooling}                         & \multicolumn{1}{c|}{Maxpooling}                         & \multicolumn{1}{c|}{{ Maxpooling}}           & Maxpooling                          \\ \hline
			\multicolumn{4}{c}{Multi-Scale Feature Aggregation}                                                                                                                                                                        \\
			\multicolumn{4}{c}{LSTM {[}128{]}}                                                                                                                                                                                         \\
			\multicolumn{4}{c}{Sample (1)}                                                                                                                                                                                             \\
			\multicolumn{4}{c}{Group (all)}                                                                                                                                                                                            \\
			\multicolumn{4}{c}{MLP {[}256,512,1024{]}}                                                                                                                                                                                 \\
			\multicolumn{4}{c}{\textbf{Architecture 1}}                                                                                                                                                                  \\
			\multicolumn{4}{c}{Maxpooling}                                                                                                                                                                                             \\
			\multicolumn{4}{c}{FC {[}512,256,40{]}}                                                                                                                                                                                    \\
			\multicolumn{4}{c}{Softmax}                                                                                                                                                                                                \\ \hline
	\end{tabular}	}
	\caption{Illustration of the Point2Sequnence with Architecture 1.}
	\label{tab:p2sw/density}
\end{table}

To obtain the Point2Sequence with Architecture 2 (as shown in Table~\ref{tab:p2sw/weight}), we added the module of Architecture 2 after the last \emph{MLP}, \emph{i.e. MLP} [256,512,1024], in Point2Sequence. This network was designed to verify the effect of Architecture 2 on the rotation robustness.

\begin{table}[htbp]
	\centering
	\resizebox{0.55\textwidth}{!}{%
		\begin{tabular}{cccc}
			\hline
			\multicolumn{4}{c}{Point2Sequence with Architecture 2}                                                                                                                                                                                   \\ \hline
			\multicolumn{4}{c}{Sample (384)}                                                                                                                                                                                           \\ \hline
			\multicolumn{1}{c|}{Group (16)}                        & \multicolumn{1}{c|}{Group (32)}                        & \multicolumn{1}{c|}{{ Group (64)}}          & Group4 (128)                        \\
			\multicolumn{1}{c|}{MLP {[}32,64,128{]}}               & \multicolumn{1}{c|}{MLP {[}64,64,128{]}}               & \multicolumn{1}{c|}{{ MLP {[}64,64,128{]}}} & MLP4 {[}128,128,128{]}              \\
			\multicolumn{1}{c|}{Maxpooling}                         & \multicolumn{1}{c|}{Maxpooling}                         & \multicolumn{1}{c|}{{ Maxpooling}}           & Maxpooling                          \\ \hline
			\multicolumn{4}{c}{Multi-Scale Feature Aggregation}                                                                                                                                                                        \\
			\multicolumn{4}{c}{LSTM {[}128{]}}                                                                                                                                                                                         \\
			\multicolumn{4}{c}{Sample (1)}                                                                                                                                                                                             \\
			\multicolumn{4}{c}{Group (all)}                                                                                                                                                                                            \\
			\multicolumn{4}{c}{MLP {[}256,512,1024{]}}                                                                                                                                                                                 \\
			\multicolumn{4}{c}{\textbf{Architecture 2}}                                                                                                                                                               \\
			\multicolumn{4}{c}{Maxpooling}                                                                                                                                                                                             \\
			\multicolumn{4}{c}{FC {[}512,256,40{]}}                                                                                                                                                                                    \\
			\multicolumn{4}{c}{Softmax}                                                                                                                                                                                                \\ \hline
		\end{tabular}%
	}
	\caption{Illustration of the Point2Sequnence with Architecture 2.}
	\label{tab:p2sw/weight}
\end{table}

To obtain the Point2Sequence with Architecture 4 (as shown in Table~\ref{tab:p2sw/oe}), we added the module of Architecture 4 after the \emph{LSTM}. This network was designed to verify the effect of Architecture 4 on the rotation robustness.

\begin{table}[htbp]
	\centering
	\resizebox{0.55\textwidth}{!}{%
		\begin{tabular}{cccc}
			\hline
			\multicolumn{4}{c}{Point2Sequence with Architecture 4}                                                                                                                                                                                   \\ \hline
			\multicolumn{4}{c}{Sample (384)}                                                                                                                                                                                           \\ \hline
			\multicolumn{1}{c|}{Group (16)}                        & \multicolumn{1}{c|}{Group (32)}                        & \multicolumn{1}{c|}{{ Group (64)}}          & Group4 (128)                        \\
			\multicolumn{1}{c|}{MLP {[}32,64,128{]}}               & \multicolumn{1}{c|}{MLP {[}64,64,128{]}}               & \multicolumn{1}{c|}{{ MLP {[}64,64,128{]}}} & MLP4 {[}128,128,128{]}              \\
			\multicolumn{1}{c|}{Maxpooling}                         & \multicolumn{1}{c|}{Maxpooling}                         & \multicolumn{1}{c|}{{ Maxpooling}}           & Maxpooling                          \\ \hline
			\multicolumn{4}{c}{Multi-Scale Feature Aggregation}                                                                                                                                                                        \\
			\multicolumn{4}{c}{LSTM {[}128{]}}                                                                                                                                                                                         \\
			\multicolumn{4}{c}{\textbf{Architecture 4 [128]}}  \\
			\multicolumn{4}{c}{Sample (1)}                                                                                                                                                                                             \\
			\multicolumn{4}{c}{Group (all)}                                                                                                                                                                                            \\
			\multicolumn{4}{c}{MLP {[}256,512,1024{]}}                                                                                                                                                                                 \\
			\multicolumn{4}{c}{Maxpooling}                                                                                                                                                                                             \\
			\multicolumn{4}{c}{FC {[}512,256,40{]}}                                                                                                                                                                                    \\
			\multicolumn{4}{c}{Softmax}                                                                                                                                                                                                \\ \hline
		\end{tabular}
	}
	\caption{Illustration of the Point2Sequnence with Architecture 4.}
	\label{tab:p2sw/oe}
\end{table}

\subsection{PointSIFT} \label{appendix:sift}

To obtain the PointSIFT without Architecture 4 (as shown in Table~\ref{tab:siftw/ooe}), we removed all the four modules of Architecture 4 from the original PointSIFT. This network was designed to verify whether Architecture 4 can improve the rotation robustness.

\begin{table}[htbp]
	\centering
	\resizebox{0.4\textwidth}{!}{
		\begin{tabular}{cc}
			\hline
			PointSIFT                               & PointSIFT without Architecture 4 \\ \hline
			\textbf{ Architecture 4 {[}64{]}}   & Sample (1024)           \\
			Sample (1024)                           & Group (0.1,32)          \\
			Group (0.1,32)                          & MLP {[}64,128{]}        \\
			MLP {[}64,128{]}                        & Maxpooling              \\
			Maxpooling                              & Sample (256)            \\
			\textbf{Architecture 4 {[}128{]}} & Group (0.2,32)          \\
			Sample (256)                            & MLP {[}128,256{]}       \\
			Group (0.2,32)                          & Maxpooling              \\
			MLP {[}128,256{]}                       & Sample (64)             \\
			Maxpooling                              & Group (0.4,32)          \\
			\textbf{Architecture 4 {[}256{]}} & MLP {[}256,512{]}       \\
			Sample (64)                             & Maxpooling              \\
			Group (0.4,32)                          & Sample (1)              \\
			MLP {[}256,512{]}                       & Group (all)             \\
			Maxpooling                              & MLP {[}512,1024{]}      \\
			\textbf{Architecture 4 {[}512{]}} & Maxpooling              \\
			Sample (1)                              & FC {[}512,256,40{]}     \\
			Group (all)                             & Softmax                 \\
			MLP {[}512,1024{]}                      &                         \\
			Maxpooling                              &                         \\
			FC {[}512,256,40{]}                     &                         \\
			Softmax                                 &                         \\ \hline
		\end{tabular}
	}	
	\caption{Illustration of the PointSIFT without Architecture 4. Here Group ({\small$r,K$}) indicates the \emph{Group layer}, which uses the ball query search to find {\small$K$} neighboring points around each sampled point within a radius {\small$r$}.}
	\label{tab:siftw/ooe}
\end{table}

\subsection{RSCNN} \label{appendix:rscnn}
In this study, we reconstructed the RSCNN \cite{liu2019relation} using four specific modules. Table~\ref{tab:rscnn1} and Table~\ref{tab:rscnn2} compare the different versions of RSCNN, including the original one, the one with Architecture 1 \cite{wu2019pointconv}, the one with Architecture 2 \cite{wu2019pointconv}, the one with Architecture 4 \cite{jiang2018pointsift}, and the one with Architecture 3 \cite{liu2018point2sequence}.

\begin{table*}[htbp]
	\centering
	\resizebox{0.8\linewidth}{!}{\begin{tabular}{c|c|c|c}
			\hline
			RSCNN             & RSCNN with Architecture 1        & RSCNN with Architecture 2        & RSCNN with Architecture 4                  \\ \hline
			
			Sample (512)           & Sample (512)                          & Sample (512)   & Sample (512)            \\
			
			Group (0.23,48) & Group (0.23,48)  & Group (0.23,48)  & Group (0.23,48)                              \\
			
			RS-Conv: MLP {[}64,16{]}    & RS-Conv: MLP {[}64,16{]}    & RS-Conv: MLP {[}64,16{]}   & RS-Conv: MLP {[}64,16{]}   \\
			
			RS-Conv: Maxpooling    & \textbf{Architecture 1}& \textbf{Architecture 2}& RS-Conv: Maxpooling\\
			
			RS-Conv: MLP {[}128{]}     & RS-Conv: Maxpooling& RS-Conv: Maxpooling & RS-Conv: MLP {[}128{]}  \\
			
			Sample (128)  & RS-Conv: MLP {[}128{]} & RS-Conv: MLP {[}128{]}  & Sample (128)                                    \\
			
			Group (0.32,64)         & Sample (128)  & Sample (128) &Group (0.32,64)   \\
			
			RS-Conv: MLP {[}32,128{]}    & Group (0.32,64) & Group (0.32,64) & RS-Conv: MLP {[}32,128{]}  \\
			
			RS-Conv: Maxpooling  & RS-Conv: MLP {[}32,128{]} & RS-Conv: MLP {[}32,128{]}  & RS-Conv: Maxpooling    \\
			
			RS-Conv: MLP {[}512{]}     & \textbf{Architecture 1} & \textbf{Architecture 2}& RS-Conv: MLP {[}512{]}  \\

			MLP {[}1024{]}              & RS-Conv: Maxpooling& RS-Conv: Maxpooling &\textbf{Architecture 4 [512]} \\
			
			Maxpooling    & RS-Conv: MLP {[}512{]}& RS-Conv: MLP {[}512{]}& MLP {[}1024{]}                                         \\
			
			FC {[}512,256,40{]} & MLP {[}1024 {]} & MLP {[}1024 {]}& Maxpooling                       \\
			
			Softmax& Maxpooling  & Maxpooling    & FC [512,256,40]                                        \\
			
			& FC [512,256,40]                                & FC [512,256,40]                                &           Softmax                                      \\

			& Softmax                               & Softmax                               &                                                 \\ \hline
	\end{tabular}}
	\caption{Different versions of RSCNN, including the original one, the one with Architecture 1, the one with Architecture 2, and the one with Architecture 4. The layerwise operation prefixed by ``RS-Conv:'' indicates that the operation is a component of the RS-Conv operation, which has been well introduced in Section~\ref{appendix:rscnn}.}
	\label{tab:rscnn1}
\end{table*}

\begin{table}[htbp]
	\centering
	\resizebox{0.75\linewidth}{!}{\begin{tabular}{c|c|c|c}
			\hline
			RSCNN             & \multicolumn{3}{c}{RSCNN with Architecture 3}                                                                                              \\ \hline
			Sample (512)           & \multicolumn{3}{c}{Sample (512)}                                                                                                           \\
			\cline{2-4}
			
			Group (0.23,48)& \textbf{Group (0.075,16)}      & \textbf{Group (0.1,32)}& Group (0.23,48)\\
			
			RS-Conv: MLP [64,16]    & \textbf{RS-Conv: MLP [64,16] } & \textbf{ RS-Conv: MLP [64,16] }   &  RS-Conv: MLP [64,16]  \\
			
			RS-Conv: Maxpooling             & \textbf{RS-Conv: Maxpooling}           & \textbf{RS-Conv: Maxpooling    }                                &  RS-Conv: Maxpooling            \\

			RS-Conv: MLP {[}128{]}    &\textbf{RS-Conv: MLP {[}128{]}}  & \textbf{RS-Conv: MLP {[}128{]}    }                      & RS-Conv: MLP {[}128{]}  \\
			\cline{2-4}
			
			Sample (128)           & \multicolumn{3}{c}{{ Sample (128)}}       \\
			
			\cline{2-4}
			
			Group (0.32,64)&\textbf{ Group (0.1,16)}      & \textbf{Group (0.15,48)}&  Group (0.32,64)   \\
			
			RS-Conv: MLP [32,128]  & \textbf{RS-Conv: MLP [32,128] }  & \textbf{RS-Conv: MLP [32,128]   }  & RS-Conv: MLP [32,128] \\
			
			RS-Conv: Maxpooling             &\textbf{RS-Conv: Maxpooling}           & \textbf{RS-Conv: Maxpooling   }   & RS-Conv: Maxpooling         \\

			RS-Conv: MLP {[}512{]}    & \textbf{RS-Conv: MLP {[}512{]}}  & \textbf{RS-Conv: MLP {[}512{]}  }                        &  RS-Conv: MLP {[}512{]}\\
			
			\cline{2-4}
			
			MLP {[}1024{]} & \multicolumn{3}{c}{{ MLP {[}1024{]} }}                                                                                      \\
			
			Maxpooling             & \multicolumn{3}{c}{{ Maxpooling}}                                                                                     \\
			
			FC {[}512,256,40{]}    & \multicolumn{3}{c}{FC {[}512,256,40{]}}
			\\
			
			Softmax                & \multicolumn{3}{c}{Softmax}
			\\ \hline
	\end{tabular}}
	\caption{The original RSCNN and the RSCNN with Architecture 3. The layerwise operation prefixed by ``RS-Conv:'' indicates that the operation is a component of the RS-Conv operation, which has been well introduced in Section~\ref{appendix:rscnn}.}
	\label{tab:rscnn2}
\end{table}

To obtain the RSCNN with Architecture 1 (as shown in Table~\ref{tab:rscnn1}), we added modules of Architecture 1 before all \emph{RS-Conv: Maxpooling} modules in the RSCNN. Architecture 1 used in this study was an MLP with two layers, the first layer contained 16 hidden units, and the second layer contained 1 hidden unit. This network was designed to verify the effect of Architecture 1 on the adversarial robustness.

To obtain the RSCNN with Architecture 2 (as shown in Table~\ref{tab:rscnn1}), we added modules of Architecture 2 before all \emph{RS-Conv: Maxpooling} modules in the RSCNN. Architecture 2 used in this study was an MLP with a single-layer, which contained 32 hidden units. This network was designed to verify the effect of Architecture 2 on the rotation robustness.

To obtain the RSCNN with Architecture 4 (as shown in Table~\ref{tab:rscnn1}), we added the module of Architecture 4 after the \emph{RS-Conv: MLP [512] layer} in the RSCNN. This network was designed to verify the effect of Architecture 4 on the rotation robustness.

To obtain the RSCNN with Architecture 3 (as shown in Table~\ref{tab:rscnn2}), we added two blocks after the first \emph{Sample layer}, \emph{i.e.} {\small$[\emph{Group} (0.075,16)\to \emph{RS-Conv: MLP} [64,16]\to \emph{RS-Conv: Maxpooling}\to \emph{RS-Conv: MLP} [128]$} and {\small$[\emph{Group} (0.1,32)\to \emph{RS-Conv: MLP} [64,16]\to \emph{RS-Conv: Maxpooling}\to \emph{RS-Conv: MLP} [128]$}. The multi-scale version added another two blocks after the second \emph{Sample layer}, \emph{i.e.} {\small$[\emph{Group} (0.1,16)\to \emph{RS-Conv: MLP} [32,128]\to \emph{RS-Conv: Maxpooling}\to \emph{RS-Conv: MLP} [512]$} and {\small$[\emph{Group} (0.15,48)\to \emph{RS-Conv: MLP} [32,128]\to \emph{RS-Conv: Maxpooling}\to \emph{RS-Conv: MLP} [512]$}. In this way, the multi-scale version of RSCNN extracted two more scale features. This network was used to verify the effect of Architecture 3 on the adversarial robustness and the neighborhood inconsistency.

\section{Implementation details about how to extend the entropy-based method \cite{ma2019quantifying}}\label{appendix:fix}
\begin{figure}[htbp]
	\centering
	\includegraphics[width=0.45\linewidth]{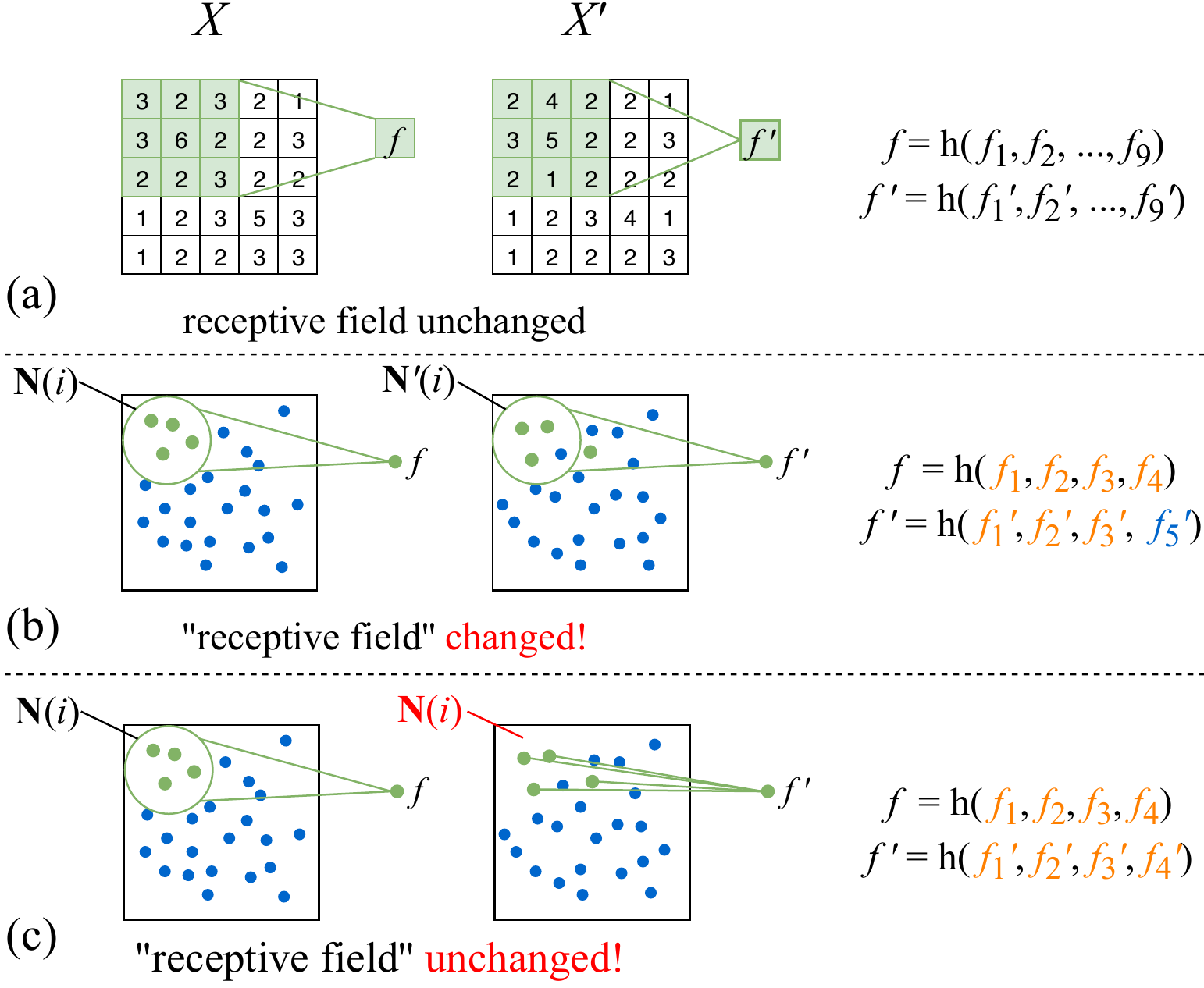}
	\caption{\textbf{Illustration of fixed sampling and grouping.} {\small$f_i$} denotes the pixel value/point-wise feature of pixel/point {\small$x_i$}.}
	\label{fig:fixedsampling}
\end{figure}

In this study, we used the entropy-based method \cite{ma2019quantifying} to quantify the layerwise information discarding of DNNs. This method assumed the feature space of the concept of a specific object satisfied {\small$\left\|f'-f\right\|^2 < \epsilon$}, where {\small$f=h(X)$}, {\small$f'=h(X')$}, {\small$X'=X+\boldsymbol{\delta}$}. {\small$\boldsymbol{\delta}$} denoted a random noise. For image processing, changing the pixel values would not change the receptive field of an interneuron, thereby features {\small$f$} and {\small$f'$} were computed using the same set of pixels (as shown in Figure~\ref{fig:fixedsampling} (a)). However, for point cloud processing, changing the coordinates of points would change the ``receptive field'' of an interneuron, \emph{i.e.} features {\small$f$} and {\small$f'$} were computed using contexts of different set of points (as shown in Figure~\ref{fig:fixedsampling} (b)).

To extend the entropy-based method to point cloud processing, we selected the same set of points as the contexts \emph{w.r.t.} {\small$x_i$} and {\small$x'_i$}. In this way, each dimension of {\small$f$} and {\small$f'$} were computed based on the same context (as shown in Figure~\ref{fig:fixedsampling} (c)). To simplify the description, here let {\small$f$} and {\small$f'$} denote local features that were computed using contextual information of {\small$x_i$} and {\small$x'_i$}, \emph{i.e.} {\small$f=h(  \{f_j|j\in \mathbf{N}(i)\})$}, and {\small$f'=h(  \{f'_j|j\in \mathbf{N}'(i)\})$}, where {\small$\mathbf{N}(i)$} and {\small$\mathbf{N}'(i)$} denoted local regions of {\small$x_i$} and {\small$x'_i$}. As shown in Figure~\ref{fig:fixedsampling} (b), changing the coordinates of points would change the ``receptive field'', \emph{i.e.} {\small$\mathbf{N}'(i)\ne \mathbf{N}(i)$}, {\small$f'$} and {\small$f$} were computed using different set of points. In order to keep the ``receptive field'' unchanged, {\small$f'$} was computed as {\small$f'=h(  \{f'_j|j\in \mathbf{N}(i)\} ) $}. In this way, features {\small$f$} and {\small$f'$} were computed using information of the same set of points.

\section{Classification accuracy comparison of different versions of DNNs}\label{appendix:accuracy}

\begin{table}[tbp]
	\newcommand{\tabincell}[2]{\begin{tabular}{@{}#1@{}}#2\end{tabular}}
	\begin{center}
		\resizebox{0.75\linewidth}{!}{
			\begin{tabular}{c|c|c|c|c|c|c|c}
				\hline
				\multirow{2}{*}{Architecture} \!\!\!&\!\!\! \multirow{2}{*}{Model} \!\!\!&\multicolumn{2}{c|}{ModelNet40}& \multicolumn{2}{c|}{ShapeNet}& \multicolumn{2}{c}{3D MNIST}\\
				\cline{3-4}\cline{5-6}\cline{7-8}
				& & w/ & w/o & w/ & w/o  & w/ & w/o  \\
				\hline
				\multirow{4}{*}{\centering \tabincell{c}{Architecture 1}}
				&PointConv& 89.02 & 88.33 & 98.50 &98.53&95.00&95.40\\
				\cline{2-8}
				&PointNet++& 90.07&89.58 &98.82&98.78& 96.10& 95.00\\
				\cline{2-8}
				&Point2Sequence&90.35  &90.84& 98.88 &98.57&99.58&93.90\\
				\cline{2-8}
				&RSCNN& 92.02 & 92.46& 98.50 &98.40 & 99.10&99.30\\
				\hline
				\multirow{4}{*}{\centering \tabincell{c}{Architecture 2}}
				&PointConv& 85.94   & 85.33 & 96.07 &96.59&85.20&89.10\\
				\cline{2-8}
				&PointNet++& 82.21& 85.65&95.82&97.13& 82.50& 87.10\\
				\cline{2-8}
				&Point2Sequence& 85.49  &88.45& 93.95 &96.63&77.30&87.09\\
				\cline{2-8}
				&RSCNN& 85.72 &86.29 & 95.86 &96.88 & 81.93 &82.17\\
				\hline
				\multirow{4}{*}{\centering \tabincell{c}{Architecture 3}}
				&PointNet++&89.50 &89.58&98.43&98.78&95.60&95.00\\
				\cline{2-8}
				&RSCNN& 92.63 &92.46 & 99.20 & 98.40 & 99.23 &99.30\\
				\cline{2-8}
				& \multicolumn{1}{c|}{Point2Sequence (4 scales \emph{vs.} 3 scales)} & \multicolumn{1}{c|}{\multirow{2}{*}{90.84}} & \multicolumn{1}{c|}{91.28} & \multicolumn{1}{l|}{\multirow{2}{*}{98.57}} & \multicolumn{1}{l|}{98.71} & \multicolumn{1}{l|}{\multirow{2}{*}{93.90}} & \multicolumn{1}{c}{94.10} \\ \cline{2-2} \cline{4-4} \cline{6-6} \cline{8-8}
				& \multicolumn{1}{l|}{Point2Sequence (4 scales \emph{vs.} 2 scales)} & \multicolumn{1}{c|}{}                       & \multicolumn{1}{l|}{91.00} & \multicolumn{1}{l|}{}                       & \multicolumn{1}{l|}{98.74} & \multicolumn{1}{l|}{}                       & \multicolumn{1}{c}{93.00} \\
				\hline
				\multirow{4}{*}{\centering \tabincell{c}{Architecture 4}}
				&PointSIFT& 83.27  &84.01 & 96.26 &91.68&84.40&90.93\\
				\cline{2-8}
				&PointNet++& 85.98  & 85.64& 94.40&97.13& 88.81&87.10\\
				\cline{2-8}
				&Point2Sequence&81.20 &88.45& 93.51&96.63&78.33&87.09\\
				\cline{2-8}
				&RSCNN& 85.03  & 86.29&97.05  & 96.88 & 83.01 &82.17\\
				\hline
		\end{tabular}}
		\caption{Accuracy of different versions of DNNs on ModelNet40, ShapeNet, and 3D MNIST. For DNNs with or without Architecture 2 and Architecture 4, during the training time, all point clouds were rotated by random angles. For DNNs with or without Architecture 1 and Architecture 3, during the training time, all point clouds were rotated around z-axis. We compared the top-1 accuracy of the network with and without each specific architecture.}
		\label{tab:acc-rr}
	\end{center}
\end{table}

Note that this study does not aim to improve the accuracy of DNNs. This study focuses on the utility analysis of different network architectures. Table~\ref{tab:acc-rr} lists the top-1 accuracy comparison results of different versions of DNNs on three different datasets, including ModelNet40, ShapeNet, and 3D MNIST.

From Table~\ref{tab:acc-rr} we can see, adding Architecture 1 had relatively equal positive and negative effects on performance. For PointConv, the network with Architecture 1 performed better than the network without Architecture 1 on the ModelNet40 dataset, while worse on the ShapeNet dataset and the 3D MNIST dataset. For PointNet++, the network with Architecture 1 performed better than the network without Architecture 1 on all three datasets. For Point2Sequence, the network with Architecture 1 performed better than the network without Architecture 1 on the ShapeNet dataset and the 3D MNIST dataset, while worse on the ModelNet40 dataset. For RSCNN, the network with Architecture 1 performed better than the network without Architecture 1 on the ShapeNet dataset, while worse on the ModelNet40 dataset and the 3D MNIST dataset.

Experimental results indicate that adding Architecture 2 to existing DNNs had negative effects on performance. For PointConv, the network with Architecture 2 performed better than the network without Architecture 2 on the ModelNet40 dataset, while worse on the ShapeNet dataset and the 3D MNIST dataset. For PointNet++, Point2Sequence, and RSCNN, networks with Architecture 2 performed worse than networks without Architecture 2 on all three datasets.

Experimental results show that adding Architecture 3 had relatively equal positive and negative effects on performance. For PointNet++, the network with Architecture 3 performed better than the network without Architecture 3 on the 3D MNIST dataset, while worse on the ModelNet40 dataset and the ShapeNet dataset. For RSCNN, the network with Architecture 3 performed better than the network without Architecture 3 on the ModelNet40 dataset and the ShapeNet dataset, while worse on the 3D MNIST dataset. For Point2Sequence, removing features extracted from neighborhoods with different scales decreased the accuracy on the ModelNet40 dataset, while increased the accuracy on the ShapeNet dataset.

Experimental results indicate that, in most cases, adding Architecture 4 to existing DNNs had negative effects on performance. For PointSIFT, networks with Architecture 4 performed better than networks without Architecture 4 on the ShapeNet dataset, while worse than networks without Architecture 4 on the ModelNet40 dataset and the 3D MNIST dataset. For PointNet++, networks with Architecture 4 performed better than networks without Architecture 4 on the ModelNet40 dataset and the 3D MNIST dataset, while worse than networks without Architecture 4 on the ModelNet40 dataset and the ShapeNet dataset. For Point2Sequence, networks with Architecture 4 performed worse than networks without Architecture 4 on all three datasets. For RSCNN, the network with Architecture 4 performed better than the network without Architecture 4 on the ShapeNet dataset and the 3D MNIST dataset, while worse on the ModelNet40 dataset.

\section{Relationship with learning interpretable representations}\label{appendix:relatedwork}

Compared to the visualization or diagnosis of representations, directly learning interpretable representations is more meaningful to improving the transparency of DNNs. In the capsule nets \cite{sabour2017dynamic,zhao20193d}, meaningful capsules, which were composed of a group of neurons, were learned to represent specific entities. \cite{vaughan2018explainable} learned explainability features with additive nature. The infoGAN \cite{chen2016infogan} learned disentangled representations for generative models. The $\beta$-VAE \cite{higgins2017beta} further developed a measure to quantitatively compare the degree of disentanglement learnt by different models. \cite{zhang2018interpretable} proposed an interpretable CNN, where filters were mainly activated by a certain object part. \cite{fortuin2018som} learned interpretable low-dimensional representations of time series and provided additional explanatory insights. \cite{mott2019towards} presented a soft attention mechanism for the reinforcement learning domain, the interpretable output of which can be used by the agent to decide its action.

\end{document}